\documentclass{article}

\usepackage{microtype}
\usepackage{graphicx}
\usepackage{subfigure}
\usepackage{booktabs} 
\usepackage{comment}
\usepackage{multirow}
\usepackage{multicol}
\usepackage{xspace}
\usepackage{url}
\usepackage{adjustbox}
\usepackage{amssymb} 
\usepackage{pifont}
\usepackage{subcaption}
\usepackage{cuted}
\usepackage{xcolor}
\usepackage{fancyhdr}

\usepackage{hyperref}

\newcommand{\squishlist}{
   \begin{list}{$\bullet$}
    { \setlength{\itemsep}{0pt}      \setlength{\parsep}{3pt}
      \setlength{\topsep}{3pt}       \setlength{\partopsep}{0pt}
      \setlength{\leftmargin}{1.0em} \setlength{\labelwidth}{1em}
      \setlength{\labelsep}{0.5em} } }
\newcommand{\squishend}{
    \end{list}  }

\usepackage[accepted]{mlsys2024}

\mlsystitlerunning{VQPy: An Object-Oriented Approach to Modern Video Analytics}

\newcommand{\us}[1]{$\mu$s}

\newcommand{\x}[1]{$\times$}
\definecolor{pink}{rgb}{1.0,0.47,0.6}
\definecolor{orange}{rgb}{1.0,0.5,0.0}
\definecolor{cyanish}{rgb}{0,0.8,1.0}
\definecolor{maroon}{rgb}{1.0,0.5,0.5}

\newcommand{\mycommand}[1]{\texttt{#1}}
\newcommand{\LogWrite}[1]{\mycommand{LogWrite}}
\newcommand{\Commit}[1]{\mycommand{Commit}}
\newcommand{\Abort}[1]{\mycommand{Abort}}
\newcommand{\WriteBack}[1]{\mycommand{WriteBack}}
\newcommand{\AtomicWrite}[1]{\mycommand{AtomicWrite}}
\newcommand{\ShortAtomicWrite}[1]{\mycommand{ShortAtomicWrite}}

\newcommand{\mystate}[1]{\textsc{#1}}
\newcommand{\Free}[1]{\mystate{Free}}
\newcommand{\Pending}[1]{\mystate{Pending}}
\newcommand{\Committed}[1]{\mystate{Committed}}

\newcommand{\Valid}[1]{\mystate{Valid}}
\newcommand{\Marked}[1]{\mystate{Marked}}

\newcommand{\eg}{\textit{e.g.}}
\newcommand{\ie}{\textit{i.e.}}

\newcommand{\etc}{\textit{etc.}\xspace}

\newcommand{\beforecaption}{\vspace{-.15cm}\begin{spacing}{0.85}}
\newcommand{\aftercaption}{\vspace{-.45cm}\end{spacing}}

\newcommand{\tool}{VQPy\xspace}

\newcommand{\query}{\codeIn{Query}\xspace}
\newcommand{\relation}{\codeIn{Relation}\xspace}
\newcommand{\vobj}{\codeIn{VObj}\xspace}
\newcommand{\vobjs}{\codeIn{VObjs}\xspace}
\newcommand{\relations}{\codeIn{Relations}\xspace}
\newcommand{\querys}{\codeIn{Queries}\xspace}

\newcommand{\fconstraint}{\codeIn{frame\_constraint}\xspace}
\newcommand{\vconstraint}{\codeIn{video\_constraint}\xspace}
\newcommand{\foutput}{\codeIn{frame\_output}\xspace}
\newcommand{\voutput}{\codeIn{video\_output}\xspace}

\definecolor{mblue}{rgb}{0.27,0.33,0.53}

\renewcommand{\em}{\it}

\newcommand{\MyPara}[1]{\noindent\textbf{#1}}

\newcommand{\codeIn}[1]{{\small\fontfamily{zi4}\selectfont{#1}}}

\newcommand{\ignore}[1]{}

\newcommand{\boldpara}[1]{\noindent{\textbf{#1}}}

\def\cffigure[#1,#2,#3]{
\begin{floatingfigure}{3.5in}
\vspace*{-2mm}
\begin{center}

\includegraphics[width=3in]{#1} 
 
\vspace*{-3mm}\caption[]{#2\vspace*{3ex}}
\label{#3} 
\vspace*{-5mm}
\end{center}
\vspace*{-2mm}
\end{floatingfigure}}

\def\cfigure[#1,#2,#3]{
\begin{figure}
\vspace*{0mm}
\begin{center}

\includegraphics[width=3in]{#1} 
 
\vspace*{-3mm}\caption[]{#2
} \label{#3}
 
\vspace*{-5mm}
\end{center}
\vspace*{-2mm}
\end{figure}}

\def\wfigure[#1,#2,#3]{
\begin{figure*}
\vspace*{0mm}
\begin{center}
 
\includegraphics[width=6in]{#1} 
 
\vspace*{-3mm}\caption[]{#2
} \label{#3}
 
\vspace*{-5mm}
\end{center}
\vspace*{-2mm}
\end{figure*}}

\def\dcfigure[#1,#2,#3,#4,#5,#6]{
{
\begin{figure*}
\vspace*{0.0in}\
\begin{center}
\begin{minipage}[c]{3in}{
\includegraphics[width=3in]{#1} 
\vspace*{-3mm}\caption[]{#2} \label{#3} \
}\end{minipage}\hspace*{0.5in}\
\begin{minipage}[c]{3in}{
\includegraphics[width=3in]{#4} 
\vspace*{-3mm}\caption[]{#5}\label{#6} \
}\end{minipage}
\end{center}
\vspace*{-0.4in}\
\end{figure*}
}
}

\def\threefigure[#1,#2,#3,#4,#5]{
\begin{figure*}
\vspace*{0mm}
\begin{center}

\begin{tabular}{ccc}
\includegraphics[width=2in]{#1} & \includegraphics[width=2in]{#2} &  \includegraphics[width=2in]{#3} \\
(a) & (b) & (c) \\
\end{tabular}

\vspace*{-3mm}\caption[]{#4
} \label{#5}

\vspace*{-5mm}
\end{center}
\vspace*{-2mm}
\end{figure*}}

\def\dssfigure[#1,#2,#3,#4,#5,#6]{
{
\begin{figure*}
\vspace*{0.2in}\
\begin{center}
\begin{minipage}[c]{4in}{
\includegraphics[width=4in]{#1}
\vspace*{-3mm}\caption[]{#2} \label{#3} \
}\end{minipage}\hspace*{0.5in}\
\begin{minipage}[c]{2in}{
\includegraphics[width=2in]{#4}
\vspace*{-3mm}\caption[]{#5}\label{#6} \
}\end{minipage}
\end{center}
\vspace*{-0.4in}\
\end{figure*}
}
}

\def\dsfigure[#1,#2,#3,#4,#5,#6]{
{
\begin{figure*}
\vspace*{0.2in}\
\begin{center}
\begin{minipage}[c]{3in}{
\includegraphics[width=3in]{#1}
\vspace*{-3mm}\caption[]{#2} \label{#3} \
}\end{minipage}\hspace*{0.5in}\
\begin{minipage}[c]{3in}{
\hspace*{0.5in}\
\includegraphics[height=3in]{#4}
\vspace*{-3mm}\caption[]{#5}\label{#6} \
}\end{minipage}
\end{center}
\vspace*{-0.4in}\
\end{figure*}
}
}

\def\dsyfigure[#1,#2,#3,#4,#5,#6]{
{
\begin{figure*}
\vspace*{0.2in}\
\begin{center}
\begin{minipage}[c]{2.5in}{
\includegraphics[height=2.5in]{#1}
\vspace*{-3mm}\caption[]{#2} \label{#3} \
}\end{minipage}\hspace*{0.5in}\
\begin{minipage}[c]{2.5in}{
\includegraphics[height=2.5in]{#4}
\vspace*{-3mm}\caption[]{#5}\label{#6} \
}\end{minipage}
\end{center}
\vspace*{-0.4in}\
\end{figure*}
}
}

\def\dyfigure[#1,#2,#3,#4,#5,#6]{
{
\begin{figure*}
\vspace*{0.2in}\
\begin{center}
\begin{minipage}[c]{3in}{
\includegraphics[height=3in]{#1} 
\vspace*{-3mm}\caption[]{#2} \label{#3} \
}\end{minipage}\hspace*{0.5in}\
\begin{minipage}[c]{3in}{
\includegraphics[height=3in]{#4} 
\vspace*{-3mm}\caption[]{#5}\label{#6} \
}\end{minipage}
\end{center}
\vspace*{-0.4in}\
\end{figure*}
}
}

\def\dyoldfigure[#1,#2,#3,#4,#5,#6]{
{
\begin{figure*}
\vspace*{0.2in}\
\begin{center}
\begin{minipage}[c]{3in}{
\epsfysize=2.0in\
\hspace{0.5in}\
\epsfbox{#1}
\vspace*{-3mm}\caption[]{#2} \label{#3} \
}\end{minipage}\hspace*{0.25in}\
\begin{minipage}[c]{3in}{
\epsfysize=2.0in\
\hspace{0.5in}\
\epsfbox{#4}
\vspace*{-3mm}\caption[]{#5}\label{#6} \
}\end{minipage}
\end{center}
\vspace*{-0.4in}\
\end{figure*}
}
}

\def\cfiguredouble[#1,#2,#3,#4]{
\begin{figure}
\vspace*{0.2in}\
\begin{center}
\begin{minipage}[c]{1.5in}{
\epsfxsize=1.5in\
\epsfbox{#1}
}\end{minipage}\hspace*{0.1in}\
\begin{minipage}[c]{1.5in}{
\epsfxsize=1.5in\
\vspace{0.1in}\epsfbox{#2}
}\end{minipage}\vspace*{-0.10in} \caption[]{#3}\label{#4}
\end{center}
\vspace*{-0.4in}\
\end{figure}
}

\def\wpfigure[#1,#2,#3,#4]{
\begin{figure*}
\vspace*{4mm}
\begin{center}

\includegraphics[width=#4]{#1} 

\vspace*{-3mm}\caption[]{#2
} \label{#3}

\vspace*{-5mm}
\end{center}
\end{figure*}}

\def\wprfigure[#1,#2,#3,#4,#5]{
\begin{figure*}
\vspace*{4mm}
\begin{center}

\includegraphics[width=#4, angle=#5]{#1} 

\vspace*{-3mm}\caption[]{#2
} \label{#3}

\vspace*{-5mm}
\end{center}
\end{figure*}}

\def\DoubleFigureWSlide[#1,#2,#3,#4,#5,#6,#7,#8,#9]{
\begin{figure*}
\vspace*{#9}
\begin{center}
\begin{minipage}{#4}
\includegraphics[width=#4]{#1}
\vspace*{-3mm}\caption{#2
}\label{#3}
\end{minipage}
\hspace{2em}
\begin{minipage}{#8}
\includegraphics[width=#8]{#5}
\vspace*{-3mm}\caption{#6
}\label{#7}
\end{minipage}
\vspace*{-5mm}
\end{center}
\end{figure*}
}

\def\DoubleFigureW[#1,#2,#3,#4,#5,#6,#7,#8]{
\begin{figure*}
\vspace*{0in}
\begin{center}
\begin{minipage}{#4}
\includegraphics[width=#4]{#1}
\vspace*{-3mm}\caption{#2
}\label{#3}
\end{minipage}
\hspace{2em}
\begin{minipage}{#8}
\includegraphics[width=#8]{#5}
\vspace*{-3mm}\caption{#6
}\label{#7}
\end{minipage}
\vspace*{-5mm}
\end{center}
\end{figure*}
}

\def\DoubleFigureWHack[#1,#2,#3,#4,#5,#6,#7,#8]{
\begin{figure*}
\vspace*{0in}
\begin{center}
\begin{minipage}{3in}
\includegraphics[width=#4]{#1}
\vspace*{-3mm}\caption{#2
}\label{#3}
\end{minipage}
\hspace{2em}
\begin{minipage}{3in}
\includegraphics[width=#8]{#5}
\vspace*{-3mm}\caption{#6
}\label{#7}
\end{minipage}
\vspace*{-5mm}
\end{center}
\end{figure*}
}

\def\ddcfigure[#1,#2,#3,#4]{
\begin{figure*}
\vspace*{0.2in}\
\begin{center}
\begin{minipage}[c]{3in}{
\includegraphics[height=3in]{#1} 
}\end{minipage}\hspace*{0.5in}\
\begin{minipage}[c]{3in}{
\includegraphics[height=3in]{#2} 
}\end{minipage}\vspace*{-0.10in} \caption[]{#3}\label{#4}
\end{center}
\vspace*{-0.4in}\
\end{figure*}
}

\def\ddcfigureSlide[#1,#2,#3,#4,#5]{
\begin{figure*}
\vspace*{#5}\
\begin{center}
\begin{minipage}[c]{3in}{
\includegraphics[height=3in]{#1} 
}\end{minipage}\hspace*{0.5in}\
\begin{minipage}[c]{3in}{
\includegraphics[height=3in]{#2} 
}\end{minipage}\vspace*{-0.10in} \caption[]{#3}\label{#4}
\end{center}
\vspace*{-0.4in}\
\end{figure*}
}

\def\cxfigure[#1,#2,#3]{
\begin{figure}
\vspace*{4mm}
\begin{center}
 
\epsfxsize=2.5in\
\epsfbox{#1}\
 
\vspace*{-0.10in}\caption[]{#2
} \label{#3}
 
\vspace*{-5mm}
\end{center}
\vspace*{-2mm}
\end{figure}}

\begin{document}
\pagestyle{fancy}
\fancyhf{}
\fancyfoot[C]{\thepage}
\pagenumbering{arabic}

\twocolumn[
\mlsystitle{VQPy: An Object-Oriented Approach to Modern Video Analytics}



\mlsyssetsymbol{equal}{*}

\begin{mlsysauthorlist}
\mlsysauthor{Shan Yu}{ucla}
\mlsysauthor{Zhenting Zhu}{ucla}
\mlsysauthor{Yu Chen}{ucla}
\mlsysauthor{Hanchen Xu}{ucla}
\mlsysauthor{Pengzhan Zhao}{ucla}
\mlsysauthor{Yang Wang}{intel}
\mlsysauthor{Arthi Padmanabhan}{hmc}
\mlsysauthor{Hugo Latapie}{cisco}
\mlsysauthor{Harry Xu}{ucla}
\end{mlsysauthorlist}

\mlsysaffiliation{ucla}{University of California, Los Angeles, California, USA}
\mlsysaffiliation{intel}{Intel, Santa Clara, California, USA}
\mlsysaffiliation{hmc}{Harvey Mudd College, Claremont, California, USA}
\mlsysaffiliation{cisco}{Cisco Research, San Jose, California, USA}

\mlsyscorrespondingauthor{Shan Yu}{shanyu1@g.ucla.edu}

\vskip 0.6in

\vspace{-4em}
\begin{abstract}
Video analytics is widely used in contemporary systems and services. At the forefront of video analytics are \emph{video queries} that users develop to find objects of particular interest. 
Building upon the insight that video objects (\eg, human, animals, cars, \etc), the center of video analytics, are similar in spirit to objects modeled by traditional object-oriented languages, we propose to develop an object-oriented approach to video analytics.
This approach, named VQPy, consists of a front-end\textemdash a Python variant with constructs that make it easy for users to express video objects and their interactions\textemdash as well as an extensible backend that can automatically construct and optimize pipelines based on video objects.
We have implemented and open-sourced VQPy, which has been productized in Cisco as part of its DeepVision framework.
\end{abstract}

]
\printAffiliationsAndNotice{}  

\section{Introduction}
\label{sec:intro}

The widespread deployment of surveillance cameras and the expansion of online video platforms have resulted in a tremendous surge in video data. Harnessing this extensive video data for intelligent video analytics is vital for practical applications such as enhancing safety in smart cities, optimizing traffic management, and enabling autonomous driving, among others.

At the core of video analytics lies the concept of video queries, which serve as a crucial link between users and video data. Video queries enable users to specify and extract video objects or events that align with their particular interests.
For instance, a traffic planner may wish to analyze patterns leading to traffic accidents, such as jaywalking or vehicles speeding past pedestrians. Likewise, a police officer might be interested in identifying suspicious activities, such as someone lingering in a restricted area or attempting to tamper with security equipment. Supporting video queries is a complex task because real-world video queries typically involve a combination of various traditional computer vision (CV) tasks, including image classification, object detection, object tracking, activity recognition, and more. Moreover, tailoring this combination of CV tasks to suit the unique query requirements of different users or applications adds an additional layer of complexity.

\MyPara{State-of-the-art.} The state-of-the-art approaches for addressing video queries can be categorized into three primary methods: (1) constructing a pipeline by hand; (2) employing a SQL-like language; and (3) using a multimodal large language model (MLLM), which provides a versatile zero-shot solution capable of handling a wide range of video queries.  Our main observation behind this work is that \emph{video queries are concerned about video objects (such as humans, animals, vehicles, \etc) and their spatial and temporal interactions}. A key limitation in these existing techniques is the lack of an \emph{object-based abstraction}, making it hard for them to both describe and optimize complex queries that center around the existence of and/or the relationship between a variety of video objects. 
We will elaborate on these approaches below; a detailed description of related work can be found in \S\ref{sec:related}.

\textit{Handcrafting pipelines.} 
The most widely adopted approach in industry for implementing a specific task is the manual crafting of pipelines that connect pretrained vision models. However, it can be a labor-intensive and error-prone process, demanding deep expertise in computer vision and significant engineering resources. In the construction of such a pipeline, CV experts are required to identify objects and analyze their relationships manually by selecting the appropriate models from model repositories (such as HuggingFace~\cite{wolf2020huggingfaces}, MMDetection~\cite{chen2019mmdetection}, or various GitHub contributions), writing inference code to query these models, and creating programs to link these tasks together.
This is a formidable undertaking and it needs to be completely redone for each new application. 

\textit{SQL-based frameworks.}
To alleviate the laborious manual efforts involved in constructing and configuring pipelines, recent video database management systems (VDBMS), exemplified by \cite{optasia, blazeit, viva, eva}, have introduced a high-level interface that permits expressive querying using a SQL-like language. This interface allows for the automatic construction of pipelines through SQL queries. However, SQL-based frameworks were originally designed for processing structured tabular data and are, therefore, not ideally suited for handling object-based video queries. This discrepancy introduces challenges in effectively expressing and executing video queries. To express complex queries that involve temporal and spatial relationships between objects (\eg, scenarios like a car approaching a cyclist), a SQL-based approach requires developers to ``think like a table''\textemdash they typically treat camera frames as if they were relational tables, with queries often involving complex nesting, joins and group-bys on these frame tables. Implementing such queries necessitates the use of many UDFs that must be coded in imperative languages like Python.

Moreover, the disparity between video objects in video queries and the structured data model used in SQL-based frameworks can result in suboptimal query optimization. 
The latter poses challenges in carrying out optimizations specifically focused on individual objects, such as memoization. For example, static properties (\eg, color) of a video object remain unchanged once computed. As demonstrated in our evaluation, remembering the values of such properties for new frames (as opposed to recomputing them) can lead to a ten-fold performance increase (\S\ref{sec:eval-sql}), while doing so in SQL is a formidable challenge.

\textit{Multimodal LLMs.} The latest development in multimodal large language models (MLLMs) empowers users to interact with and comprehend videos using natural language queries. This approach offers a solution that appears to eliminate the necessity for constructing pipelines.
While MLLMs show great promise in video understanding, they excel most in the realm of exploratory video analytics, where human users engage with the video query system to iteratively explore and gain a deeper understanding of the video content. They are unable to answer questions regarding video objects on a specific frame.
Moreover, MLLMs come with a high computational cost, which can result in substantial delays. This, combined with the iterative process of generating responses, often renders the latency too long to be suitable for time-sensitive applications like real-time surveillance.

\MyPara{Insight.} Our key insight is video objects are fundamentally similar to the objects modeled in traditional object-oriented programming languages like Java or Python. Consequently, the creation of a video-object-oriented query framework allows for the straightforward development of complex queries.
Moreover, the video-object-oriented design enables optimizations \emph{at the object level}, which can greatly enhance query performance.

\MyPara{VQPy.} The original Python language lacks built-in support for representing object interactions, spatial and temporal relationships, and constraints on video frames. As a solution, we have created \tool, which is a Python variant featuring constructs specifically tailored for expressing and modeling video objects and their relationships. \tool places video objects at the core of all queries. These video objects exhibit inheritance relationships and possess properties and operations, much like those in a conventional object-oriented (OO) language.
Basing \tool off Python is intentional to tap into the extensive machine learning ecosystem, making it seamless for developers to construct end-to-end pipelines, ranging from queries to subsequent tasks. 

\tool makes the following contributions:
\squishlist
    \item \textbf{A video-object-oriented frontend:} \tool employs an object-oriented approach to represent video objects and their interactions, enabling developers to create intricate queries with ease, without the need for additional code to connect various CV tasks or express object interactions through a relational data model for SQL. Additionally, our design embraces object-oriented concepts like inheritance and polymorphism, promoting code reusability and facilitating query composition.
    \item \textbf{An efficient backend with an object-centric data model:} Our backend is constructed based on a data model that revolves around video objects rather than relational tables. This approach streamlines the efficient execution of video-object-oriented queries, enabling the integration of numerous object-level computation reuses that were not feasible in a SQL-based framework primarily geared toward relational data.
    \item \textbf{An extensible optimization framework:} \tool's optimization engine has been crafted as a flexible optimization framework, making it effortless to integrate diverse query optimizations like frame filtering and specialized neural networks (NNs) as plug-and-play operators, requiring minimal adjustments to the code.
\squishend

\MyPara{Results.} We have written 14 queries with \tool and evaluated on 5 datasets from real-word surveillance video streams. On average, \tool achieved more than 10$\times$ query speedup over state-of-the-art systems without sacrificing accuracy. We also implemented three optimizations commonly used in previous works showcasing \tool's ability to integrate customized optimizations. \tool has been open-sourced at \emph{https://github.com/vqpy/vqpy} and integrated into DeepVision, which is Cisco's comprehensive video analytics framework, for commercial purposes.
\section{\tool Overview}
\label{sec:overview}

\begin{figure}[ht]
    \centering
    \includegraphics[width=0.95\linewidth]{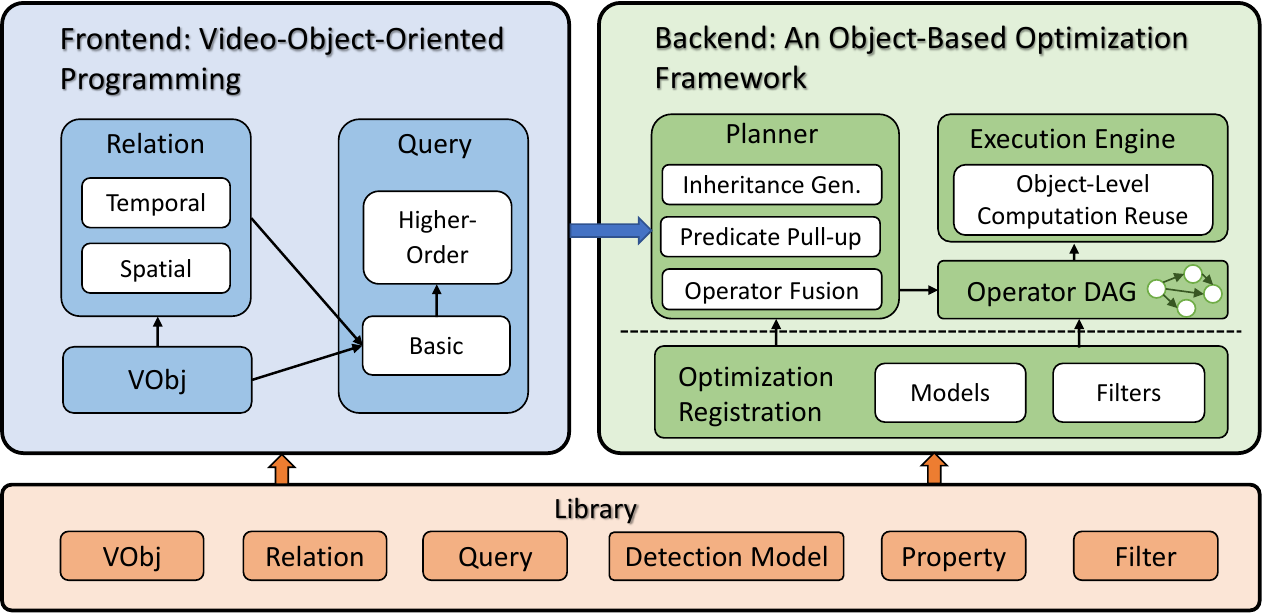}
    \vspace{-1em}
    \caption{\tool architecture.}
    \vspace{-1em}
    \label{fig:fig-overview}
\end{figure}

As shown in Figure \ref{fig:fig-overview}, \tool's architecture contains three core components: frontend, backend, and library.

\MyPara{Frontend.}
The frontend of \tool augments Python's capacity to articulate video queries through the introduction of three key constructs: \vobj, \relation, and \query. \vobj serves as the central abstraction within \tool, defining the primary objects of interest within video data. \relation builds upon \vobj, specifying spatial or temporal relationships among these objects, while \query further extends the concepts of \vobj and \relation to define a comprehensive video query.

\MyPara{Backend.}
\tool's backend framework uses an object-based data model, and comprises three fundamental components: operators, planner, and execution engine. When presented with a video query, our planner constructs a sequence of operators, such as object detection and object tracking, using a graph-based data model centered around \vobj. Subsequently, the execution engine carries out the execution of this pipeline. Moreover, \tool's backend simplifies the process of optimization registration, allowing users to effortlessly incorporate their filters and specialized neural networks for video objects into our backend through straightforward Python annotations. The planner, equipped with metadata from the property library and profiling data from the model zoo, then reorganizes the operators within the directed acyclic graph (DAG) to generate an optimized query plan that aligns with the users' specified accuracy targets.

\textbf{Library.}
 \tool provides a library that encompasses a model zoo, which integrates state-of-the-art models both for common CV tasks including object detection, action recognition, and object tracking, and for specific property functions like license plate recognition or color detection. The models in the library can be selected to construct \vobj and \relation. \tool's library also provides commonly used \codeIn{VObjs}, \codeIn{Relations} and \codeIn{Queries} that serve as building blocks for constructing other queries. Besides, \tool's library also includes backend optimizations including specialized NNs and filters corresponding to the built-in \vobjs and \querys.
\section{Frontend: Video-Object-Oriented Programming}
\label{sec:frontend}

This section discusses \tool's frontend, with a focus on how users can express queries in an object-oriented manner.
As discussed in \S\ref{sec:overview}, to empower Python with video query abilities, \tool extends Python's syntax with three major constructs:  \vobj,  \relation, and \query. These constructs are similar to Python classes but carry special properties and constraints to ease the development of video analytics. 

\MyPara{VObj.}
Much like an object in vanilla Python, \vobj defines the video object type users want to query on (\eg, vehicle, person,  \etc).
\vobj supports the definition of \emph{properties}, \eg, the color of the car, which can be used in \query for detecting objects with such properties, such as a red car. 

\begin{figure}[htbp]
\centering
\vspace{-1em}
\includegraphics[width=\linewidth]{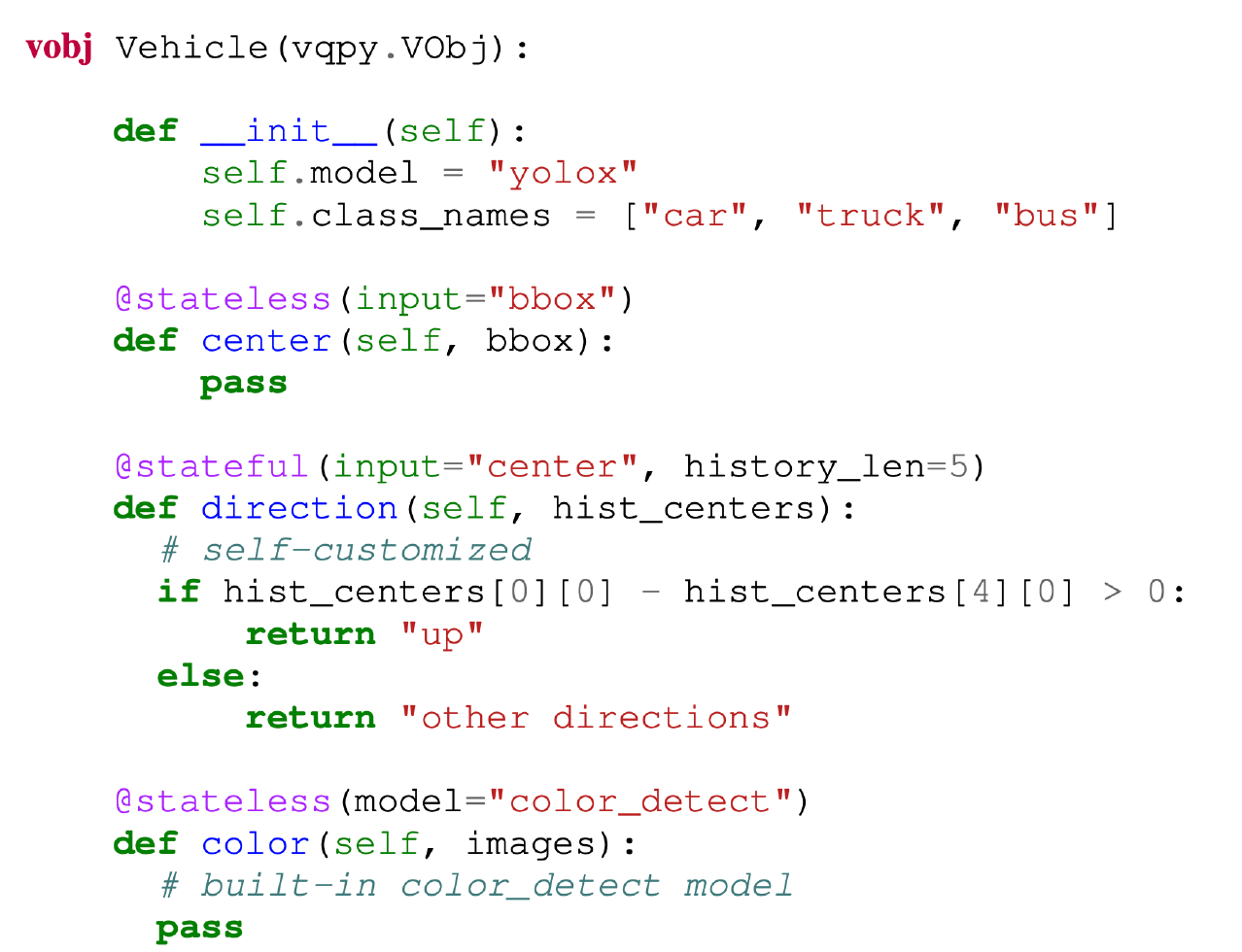}
\vspace{-1.5em}
\caption{\tool Vehicle \vobj. 
\label{fig:vobj-vehicle} \vspace{-1em}}
\end{figure}

Figure \ref{fig:vobj-vehicle} demonstrates how to construct a vehicle \vobj with the properties ``center'', ``direction'', and ``color''. In \vobj, properties can be built with other properties defined in the same \vobj (\eg, the ``direction'' property takes as input the ``center'' property of the vehicle \vobj), or with pre-defined properties in \codeIn{vqpy.VObj}  (\eg, \codeIn{bbox}, \codeIn{frame\_rate}, \codeIn{vobj\_image}, \etc), or with properties in its super-\codeIn{VObjs}.

In \vobj, each property can be either \emph{stateless} or \emph{stateful}, indicating whether the property requires cross-frame information (\eg, \codeIn{direction}) or not (\eg, \codeIn{color}). 
If a property is stateless, that is, the property is static in the \vobj and only depends on the current frame (such as ``color'' and ``license''), users can modify the property with a \codeIn{@stateless} annotation, which allows the developer to specify other properties within the same frame as dependencies. 
Similarly, to compute a stateful property, users can annotate it with a \codeIn{@stateful} annotation, which takes as input the length of the history of its dependent property. In the Vehicle \vobj example shown in Figure~\ref{fig:vobj-vehicle}, the \codeIn{direction} property of a car is a stateful property, and computing it requires five consecutive frames of the \codeIn{center} property. 

To build a \vobj in \tool, users can directly utilize the vision models from \tool's library by referring to the model name. For example, the \codeIn{Vehicle} \vobj uses the built-in ``yolox'' as its model to detect vehicle objects, and the ``color\_detect'' model to compute the color property. Besides, developers 
 can also write customized code to define properties, such as the \codeIn{direction} property in Figure~\ref{fig:vobj-vehicle}.

A special \vobj we provide is the scene \vobj, which represents the scene of each frame. This can be used to define background properties, such as day or night, rainy or sunny, whether at an intersection, \etc 

\begin{figure}[htbp]
\includegraphics[width=\linewidth]{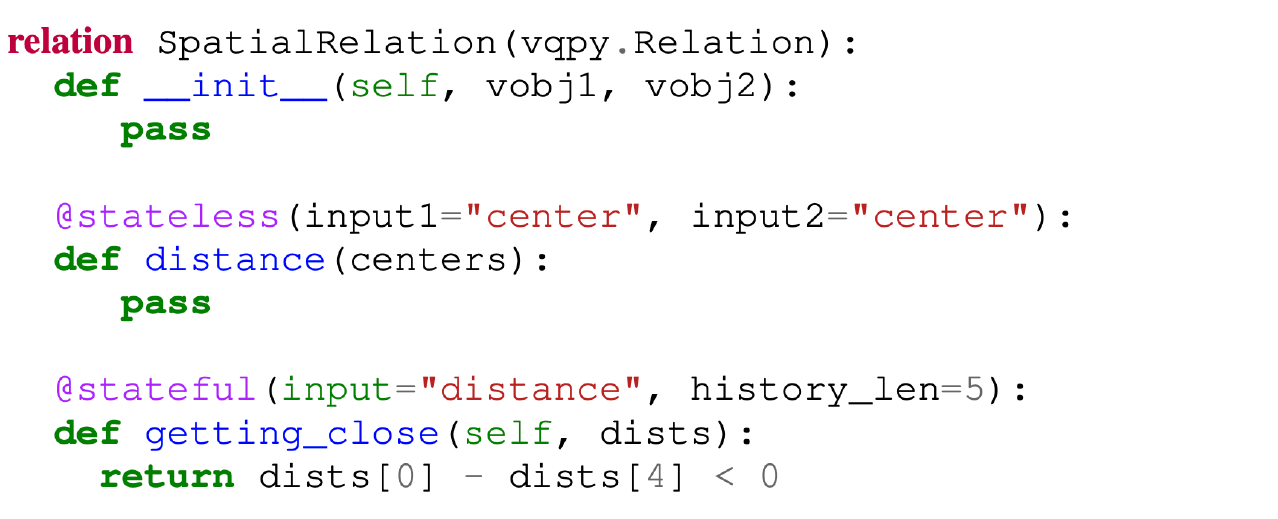}
\vspace{-1.5em}
\caption{\tool spatial relation. 
\label{fig:relation-spatial} \vspace{-1em}}
\end{figure}

\MyPara{Relation.} 
To ease the query development for object interactions, we introduce the \relation construct. Taking \vobjs as input, \relation models spatial or temporal relations between the input \vobjs. Similarly to properties on \vobj, properties on \relation can be either stateful or stateless. Figure~\ref{fig:relation-spatial} demonstrates how to use \relation to construct a spatial relation between objects using simple python code.

\begin{figure}[htbp]
\includegraphics[width=\linewidth]{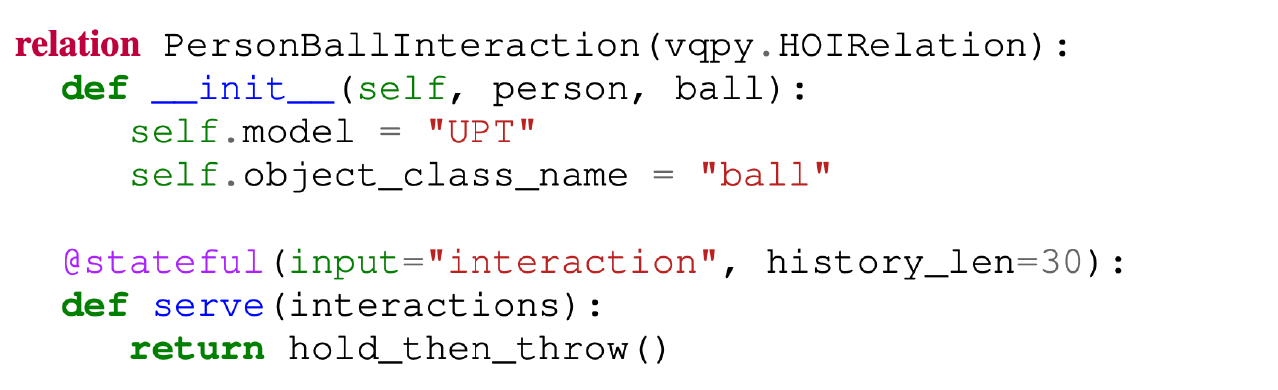}
\vspace{-1.5em}
\caption{\tool person ball relation.
\label{fig:relation-hoi} \vspace{-1em}}
\end{figure}

Instead of hand-written python code, one could also build a property with a vision model.  
For example, in Figure \ref{fig:relation-hoi}, the \codeIn{PersonBallInteraction} \relation uses the built-in human object detection model of ``UPT'' to compute the interactions between human and ball. The ``interaction'' property on \relation \codeIn{PersonBallInteraction} builds a connection between a person and a ball. 
To construct such properties, users can select vision models from \tool's library that can directly predict these properties on frames. 

\begin{figure}[htbp]
\includegraphics[width=\linewidth]{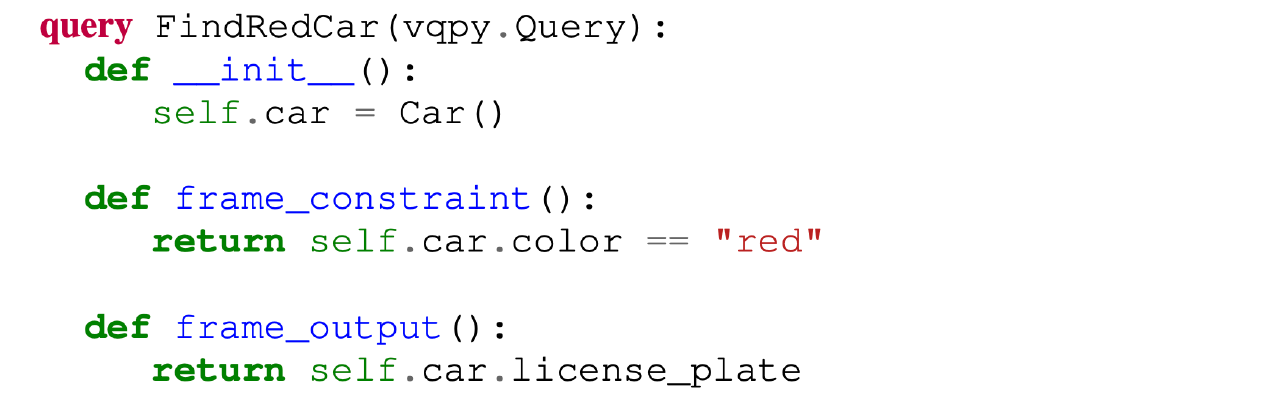}
\vspace{-1.5em}
\caption{\tool query for retrieving license plates of red cars. 
\label{fig:red-car} \vspace{-1em}}
\end{figure}

\begin{figure}[htbp]
\includegraphics[width=\linewidth]{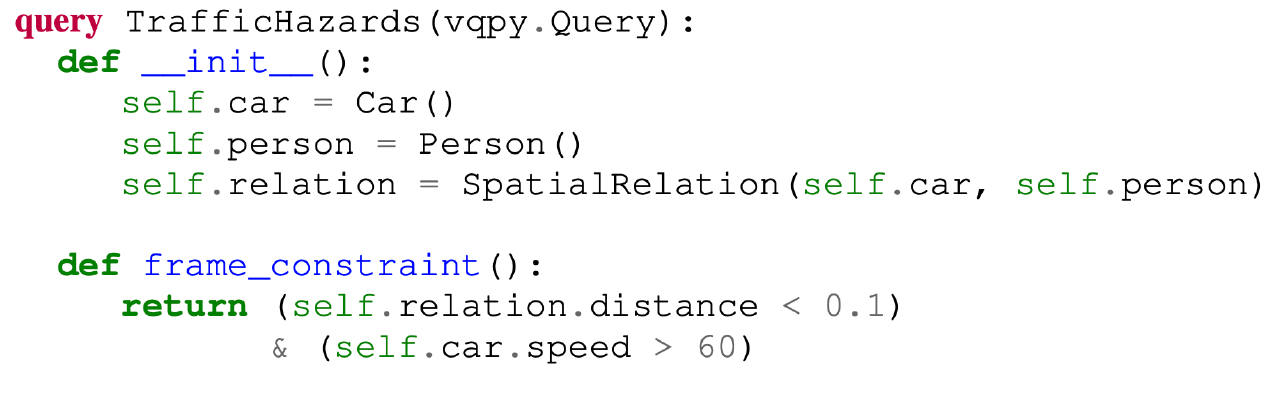}
\vspace{-1.5em}
\caption{\tool query for traffic hazards.
\label{fig:traffic-hazard} \vspace{-1em}}
\end{figure}

\MyPara{Query.}
\query is the main entry of a video query. 
With the \query construct, our goal is to enable query expressions to be semantically aligned with users' interests in video objects and their relationships.
Figure \ref{fig:red-car} demonstrates how a police officer can construct a query of ``retrieving the license plates of red cars'' with \tool.
Figure \ref{fig:traffic-hazard} includes a more complex query on both video objects (a speeding car) and their spatial relationships (a car close to a person), where a city safety guard wants to query the traffic hazard case of a speeding car passing a person. 

As shown in both examples, \tool introduces two constructs, \fconstraint and \foutput. \fconstraint allows users to express their \emph{filtering constraints} on video frames, while \foutput selects the output objects of interest. In particular, \tool invokes \foutput to output a set of video objects whose containing frames satisfy the constraints declared in \fconstraint.

\begin{figure}[htbp]
\includegraphics[width=\linewidth]{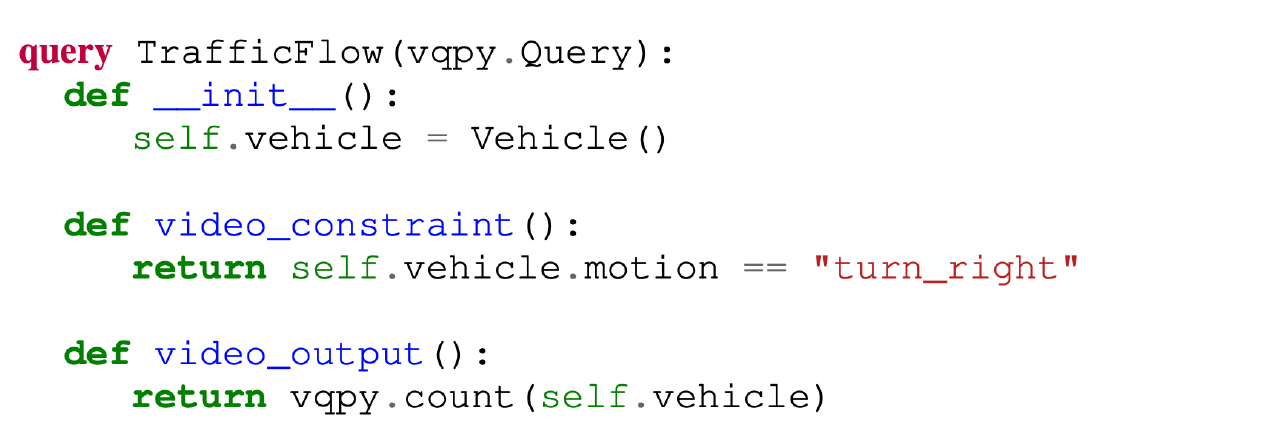}
\vspace{-1.5em}
\caption{\tool query for traffic flow analysis. 
\label{fig:traffic-flow} \vspace{-1em}}
\end{figure}

Additionally, to enable queries over the \emph{entire video}, we introduce \vconstraint and \voutput. Figure \ref{fig:traffic-flow} depicts how to express a query of ``counting the number of vehicles turning right throughout the video''. 
A \query with \vconstraint and \voutput outputs the aggregated results, where the same object that appears in different frames will be regarded as one single entity.

Note that \tool supports the use of logical operators (\&, $|$, and $\neg$) to connect the predicates in a constraint. With logical operators, queries on video objects with the conjunction or disjunction of a number of predicates (\eg, a person who wears jeans and whose hair is not black), as well as co-curing objects (\eg, frames with a person and a red car) can all be easily expressed.

\MyPara{Inheritance.} \tool supports inheritance for \vobj, \relation, and \query similar to standard python semantics. For example, a sub-\vobj or sub-\relation can inherit a super-\vobj/super-\relation, and thus all properties defined in the super-\vobj/super-\relation are directly accessible in the sub-\vobj/sub-\relation.  A sub-\query can reuse the constraints of all its super-\query to construct a stricter constraint. Inheritance facilitates code reuse and allow \tool to provide a library of basic \vobj, \relation and \query for users to extend. Inheritance also provides a natural way to enable optimizations such as specialized NNs and frame filters, which we will detail in \S\ref{sec:backend-ext}. 

\MyPara{Event Composition with Higher-Order Queries.}
Event composition allows users to express complicated queries by connecting basic queries. 
To support composition, \tool provides higher-order queries that take other queries as input and extend the query dimension temporally or spatially. Specifically, \tool offers three high-order queries: \codeIn{DurationQuery}, \codeIn{SpatialQuery}, and \codeIn{TemporalQuery}.

 A \codeIn{DurationQuery} checks whether a condition defined in the base query continues to hold for a number of frames or a time period.
 It can express queries such as a person loitering for more than 20 mins, or a bag unattended for more than 5 mins.
 A \codeIn{SpatialQuery} takes in two basic \querys, each containing a \fconstraint\ and a specific spatial relation.  \tool automatically generates a new \fconstraint\ for the  \codeIn{SpatialQuery} that checks whether the two video objects satisfy the specified spatial relationship. 
A \codeIn{TemporalQuery} takes in two \querys, each containing a \fconstraint or \vconstraint as well as a temporal relation. \tool generates a \vconstraint\ for the \codeIn{TemporalQuery} that checks whether the two events satisfy the specified temporal relationship. 

Query composition follows the following rules: \\
\textbf{Rule 1}: \codeIn{SpatialQuery} takes in only basic queries;\\
\textbf{Rule 2}: \codeIn{DurationQuery} takes in basic queries or \codeIn{SpatialQueries};\\
\textbf{Rule 3}: \codeIn{TemporalQuery} takes in basic queries as well as all three higher-order queries (including itself).

\begin{figure}[ht]
\includegraphics[width=\linewidth]{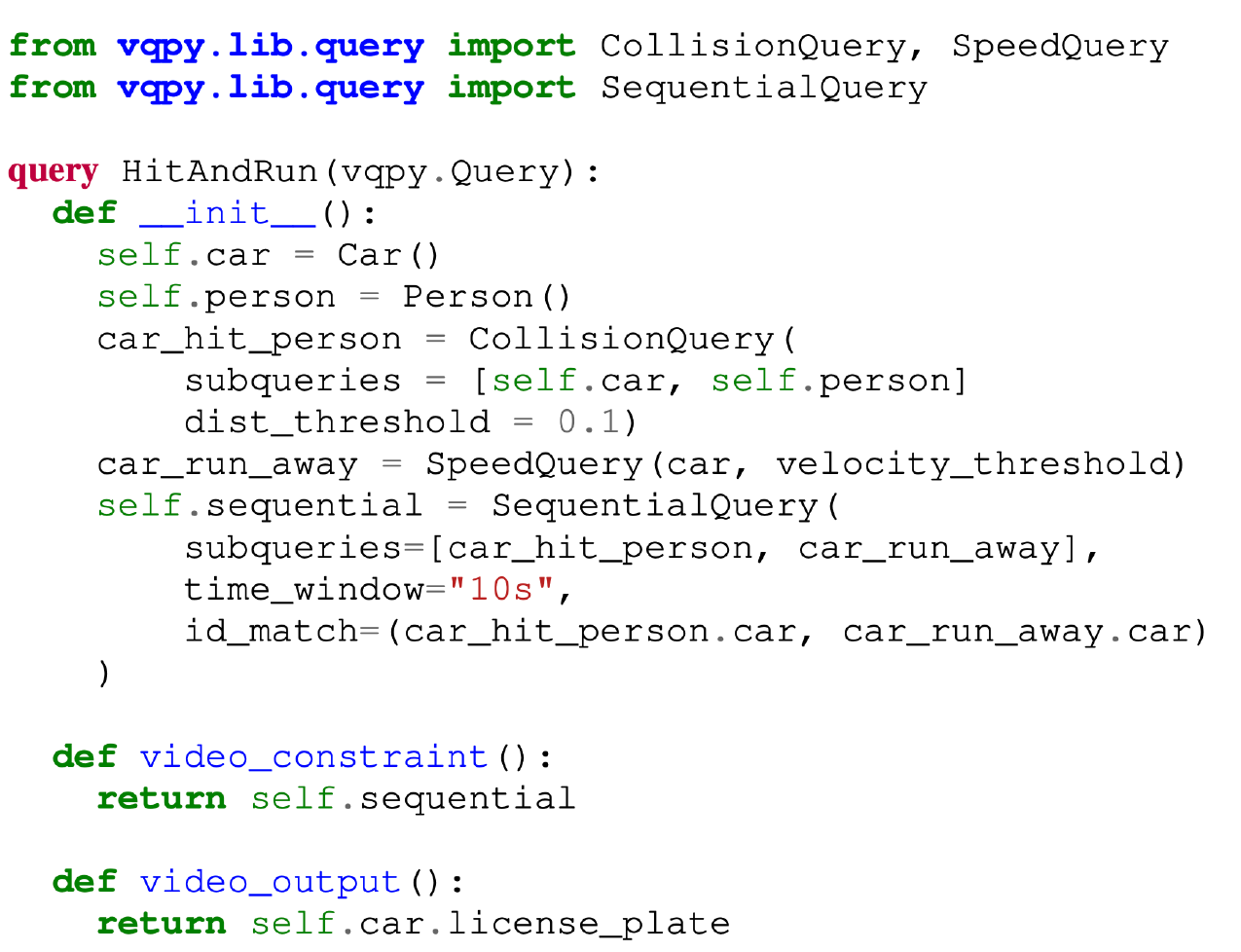}
\vspace{-1.5em}
\caption{\tool code for hit and run. 
\label{fig:hit} \vspace{-1em}}
\end{figure}

Figure~\ref{fig:hit} depicts how a traffic safety analyst can use the higher-order query constructs of \tool to implement a complicated query that searches for hit-and-run scenarios. The example includes two events, \codeIn{car-hit-person}, and \codeIn{car-run-away}, which happen sequentially. The \codeIn{car-hit-person} query is concerned with the spatial relationship between a car object and a person object. The \codeIn{car-run-away} query is interested in a car object with a cross-frame property ("speed").

In the \codeIn{HitAndRun} example, 
the user first employs \codeIn{CollisionQuery}, a sub-\query of the higher-order \codeIn{SpatialQuery} which checks whether the distance of the two input \vobjs is smaller than a threshold indicating a potential collision, for building a car-hit-person query. 

To construct this query, developers directly pass two \vobjs Car and Person.
They can then leverage \tool's built-in \codeIn{SpeedQuery}, to construct a \codeIn{car\_run\_away} query, by specifying the Car \vobj\ with a speed that exceeds \codeIn{velocity\_threshold}. 
With the two sub-queries defined, they can easily compose them into a \codeIn{SequentialQuery}, a sub-\query of the higher-order \codeIn{TemporalQuery}, by specifying a desired time window, which represents the maximum interval between the two events. The sequential query can used directly in the \vconstraint of the \codeIn{HitAndRun} query.
\section{Backend: A video-object-centric optimization framework}
\label{sec:backend}

\subsection{Plan Generation}
\label{sec:backend-plan}

As shown in Figure~\ref{fig:fig-overview}, \tool's backend includes three components, \ie, operators, query planner, and execution engine. The backend is designed with \emph{\vobjs and their relations} as its data model. 

\MyPara{Data Model.} Our query planner maintains a \emph{graph} data structure, which flows across the operators on a DAG. Nodes in the graph represent \vobjs and edges represent their relationships. There are four kinds of edges: (1) a motion edge connects two \vobjs which represent an identical object from consecutive frames (to track stateful properties such as actions), (2) a spatial-relation edge connects two \vobjs that are located in the same frame, (3) a duration-relation edge connects two \vobjs in frames whose distance is within a given time constraint, and (4) a temporal-relation edge connects two \vobjs such that the from-\vobj is in a frame that precedes the frame that contains the to-\vobj. The last three kinds of edges correspond to the three higher-order query types discussed earlier.
Nodes and edges all carry properties.

\boldpara{Operators.} \tool supports queries with six types of operators: \codeIn{video reader}, \codeIn{frame filter}, \codeIn{object detector}, \codeIn{objector tracker}, \codeIn{object filter}, and \codeIn{projector}. \codeIn{Object filter} includes \vobj\ filter and \relation\ filter;  \codeIn{projector} includes \vobj\ projector and \relation\ projector. 
\codeIn{Video reader} reads in the video stream and passes the frame information to subsequent operators. \codeIn{Frame filter} filters out irrelevant frames. For instance, a motion detector that filters out static frames can serve as a frame filter in the pipeline for answering queries including moving video objects; a texture-based filter can quickly eliminate frames that do not contain live objects such as humans or animals. \codeIn{Object filter} filters outs \vobjs or \relations\ that do not satisfy the user specified constraints. 

\codeIn{Object detector} detects video objects and labels their classes. \codeIn{Object tracker} identifies the same video object across different frames. Note that object tracker is used only when the query constraints involve \codeIn{stateful} properties.
\codeIn{Projector} outputs the results for \vobj\ or \relation\ involved in the query.

Operators are implemented as iterators. Each operator consumes the graph(s) passed from its previous operator, and outputs a new graph with updated nodes and/or edges. 
Specifically,  \codeIn{object detector} generates new nodes into the graph. Both \codeIn{object tracker} and \codeIn{projector} update edges. For example, \codeIn{object tracker} adds motion edges between existing \vobjs, and annotates them with \codeIn{track\_id}. \codeIn{Object filter} removes nodes, spatial-relation edges, or temporal-relation edges that fail the user-specified constraints. 

To support the computation of stateful properties on \vobj\ or \relation, which requires the history data of their dependent properties, the stateful projector maintains a local sliding window of historical data of all of its dependencies. When executing the DAG, the executor generates frame batches (\ie, the size of each batch is user-defined) and executes the pipeline on a per-batch basis. 

\begin{figure}[t]
    \centering
    \includegraphics[width=0.9\linewidth]{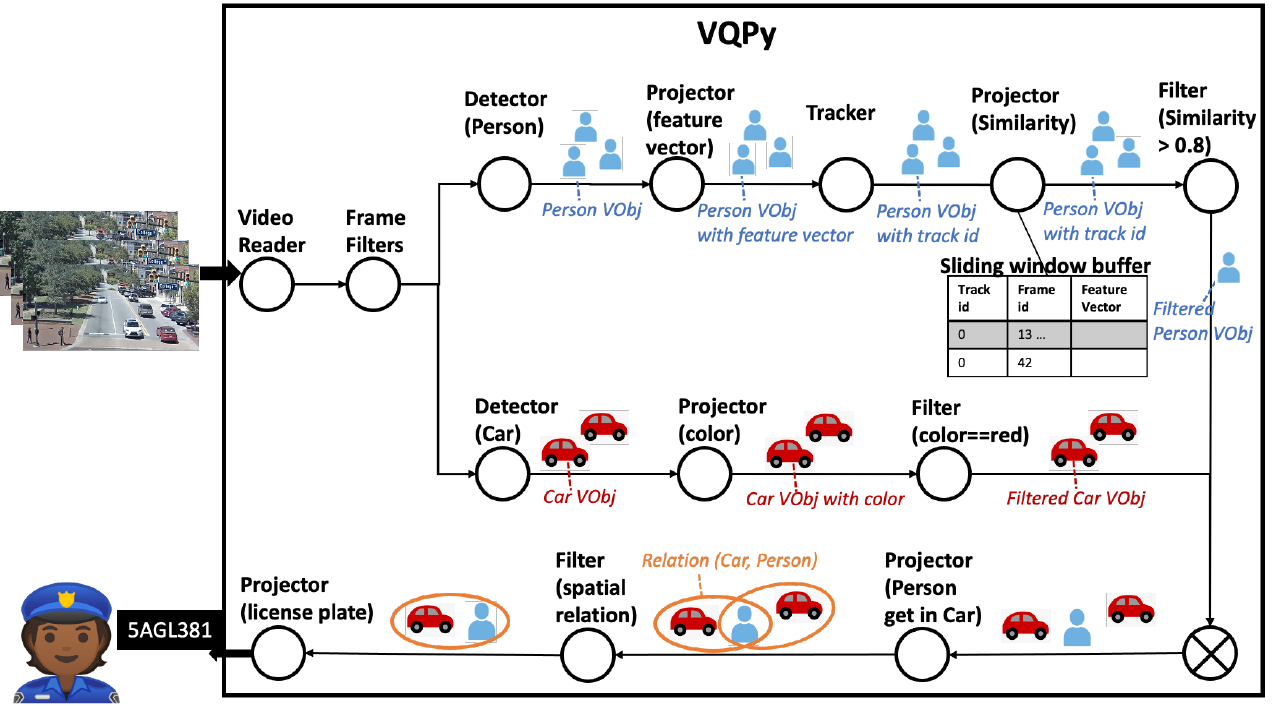}
    \vspace{-4mm}
    \caption{\tool operator DAG for a query that searches for the suspect getting in a red car.}
    \vspace{-6mm}
    \label{fig:fig-dag}
\end{figure}

\begin{figure}[ht]
\includegraphics[width=\linewidth]{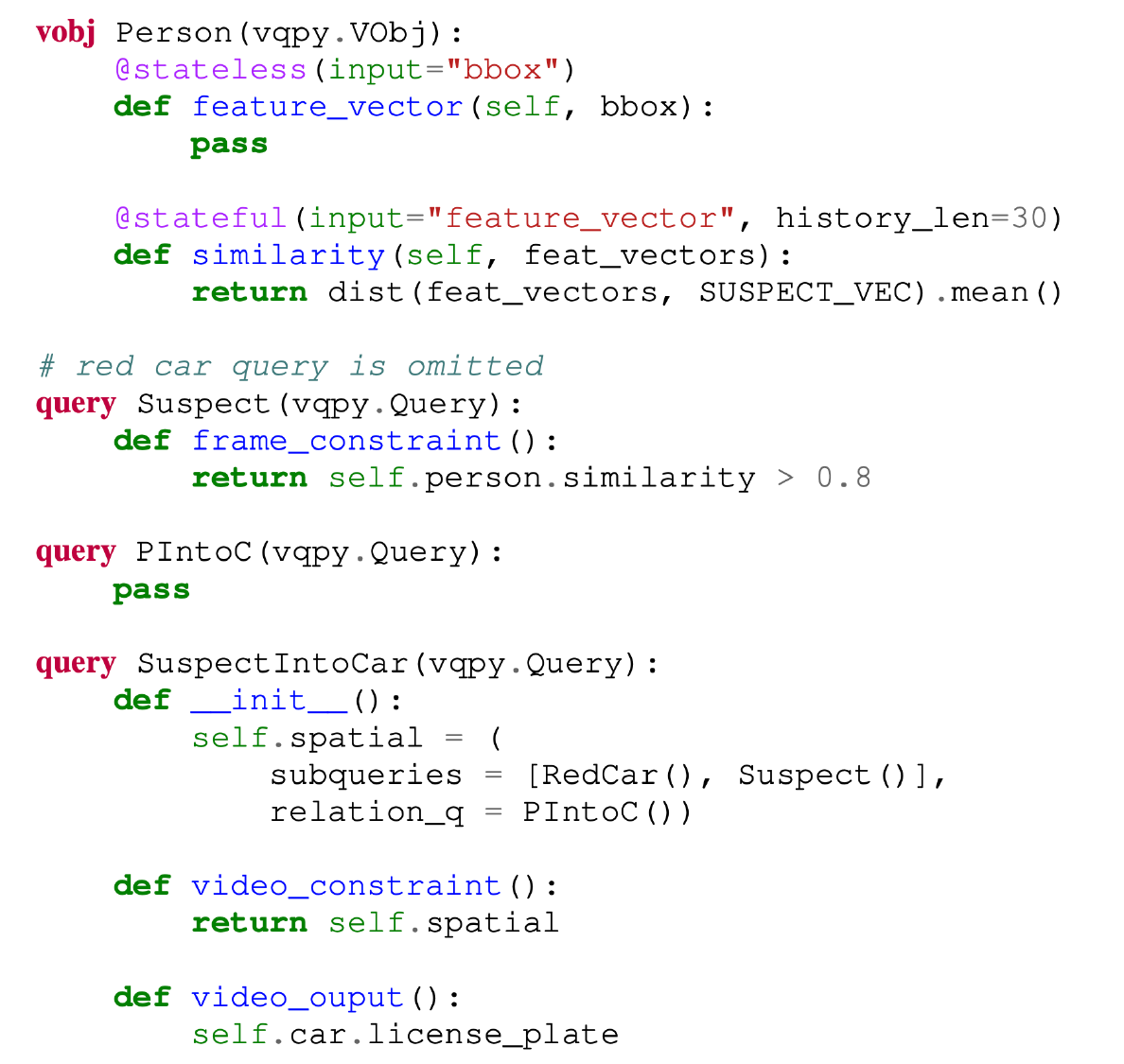}
\vspace{-1.5em}
\caption{\tool code for suspect getting in a red car. 
\label{fig:suspect} \vspace{-1em}}
\end{figure}

\MyPara{Example DAG.} Figure~\ref{fig:fig-dag} shows how a pipeline DAG is constructed and executed to answer the query that searches for a suspect getting into a red car, and identifies the license plate of the car. The code for the query is illustrated in Figure~\ref{fig:suspect}. 

A police officer uses the image of the suspect to find the person in the video.
The example query has a spatial relation query of ``person getting into car'' (\codeIn{PIntoC}), and two basic queries of a suspect person (\codeIn{Suspect}) and a red car (\codeIn{RedCar}), respectively. To find the suspect, our planner uses a human object detector that generates a graph with a single \codeIn{Person} \vobj, with two fields, a stateless \codeIn{feature\_vector} property that computes the feature vector of a person from each image, and a stateful \codeIn{similarity} property that takes in the past 30 frames of the \codeIn{feature\_vector} property, to compare the distance between each person's feature vectors and those of the target suspect and determine whether they are the same person. To identify red cars, the planner uses a car detector that generates another graph with a single \codeIn{Car} \vobj containing a ``color'' property.

To construct the DAG, \tool's planner first retrieves the dependencies of the nested query. The \codeIn{PIntoC} \query depends on the \codeIn{Suspect} and \codeIn{RedCar} \querys, and therefore the planner places all the filters and projectors related to \codeIn{PIntoC} after \codeIn{Suspect} and \codeIn{RedCar}. \codeIn{Suspect} and \codeIn{RedCar} have no dependencies between each other, so they can run in parallel. 
Next, the planner generates the detectors, trackers, projectors and filters for each query, according to the \vobjs and \relations included in the frame/video constraints. Note that multiple projectors could be generated for one predicate,  due to the dependencies between properties. In the example, two projectors for \codeIn{feature\_vector} and \codeIn{similarity} property are generated for the \codeIn{similarity > 0.8} predicate. 

The planner generates multiple frame filters and places them before the computation-expensive detectors (car detector and person detector), including frame filters based on frame difference (\eg, motion detector), and cheap filters corresponding to detectors (\eg, texture-based car filter and person filter). Note that the \codeIn{color==red} filter on the \codeIn{car} path and the \codeIn{similarity > 0.8} filter on the \codeIn{person} path are \vobj filters, that only filter out \vobjs that falsify the constraints on properties, and cannot filter out frames. The join operator serves as a frame filter that filters out the frames without \codeIn{RedCar} or \codeIn{Suspect}, merges the information of person and car retrieved from the previous operators including computed properties and filtered \vobj ids, and passes the graph with three (car and person) nodes to the spatial relation projector, which eventually adds two edges between these nodes. 

Operators can be placed on different devices. For example, the compute-intensive object detector can be placed on a GPU server while the low-cost object filter can be placed on an edge device (such as a camera). This design can easily support both offline batch and real-time streaming analytics.

\subsection{Object-level Computation Reuse}
\label{sec:backend-reuse}

The object-based data model used in our backend facilitates  \textit{object-level} computation reuse. 
In \tool, each video object (\vobj) represents a unique entity that appears across multiple frames, holding stateful and stateless properties. Reuse opportunities arise when a \vobj possesses \emph{intrinsic properties}, a special type of stateless property that remains constant across frames. Intrinsic properties are common in video queries\textemdash for example, an amber alert query may search for a red car with a license plate ending at ``45'' where both the red color and the license plate are intrinsic properties. \tool enables users to annotate stateless properties as intrinsic using \codeIn{intrinsic=True}, facilitating computation reuse at the video object level.

\tool's backend tags each video object with a label indicating whether its intrinsic properties satisfy the query constraints. Due to the static nature of these properties, this label would never change once computed. 
When processing a new frame, \tool uses a lightweight tracker based on the Kalman filter to identify \vobjs on the frame. If a \vobj has been detected before, \tool directly utilizes the \vobj's intrinsic label for sub-queries that involve intrinsic properties while only sending newly-detected \vobjs to the full computation pipeline. This optimization often leads to significantly improved computation efficiency. 

In SQL-like video query frameworks, achieving such optimizations is unattainable because of the limitations imposed by their tabular data model. Under this model, each row in the tabular structure is treated as a separate entity, rendering the task of grouping rows based on objects challenging.
Specifically, SQL frameworks lack a built-in concept of ``objects", making it impossible to implement memoization strategies at the individual object level.

\tool also supports query-level computation reuse where results from previous queries are materialized and reused when multiple queries are conducted on the same video, further improving efficiency.

\subsection{DAG Optimization}
Our planner performs three kinds of optimizations on the generated DAG: (1) operator fusion that fuses neighbor operators to reduce the overhead of executing each operator separately and minimize the intermediate data generation, (2) predicate pull-up that pulls filters to an early point of the pipeline, thereby reducing the amount of data that needs to be processed by the subsequent operators and saving computation costs, and (3) generating and comparing alternative optimization paths based on the \emph{inheritance} relationships between video objects. Since the first two optimizations were used in prior systems such as ~\cite{eva} and ~\cite{niijima-sosp19}, this subsection will focus specifically on the third optimization, which is a unique contribution of \tool. 
To realize (3), our plan generates all possible execution DAGs from a given query, each corresponding to a potential execution pipeline. The planner then profiles each DAG using a short canary input video provided by the user \textemdash a technique also employed by systems like \cite{romero}. 
The profiling helps compare the costs and accuracies of each DAG. During this process, we also identify the cost of each operator in each DAG, which can be used to perform intra-DAG optimizations such as (1) and (2).  The planner selects the best plan that meets the target accuracy with the lowest cost (best runtime). This plan can be saved for future queries on similar datasets to save optimization time. 

DAG optimizations require estimation of accuracy and performance.
For accuracy estimation, we use common techniques from recent work~\cite{blazeit, miris, thia} that uses the original models to generate ground-truth labels. We use F1 score to
estimate accuracy and compute an F1 score per DAG. To estimate each DAG’s accuracy, \tool runs the DAG with the most general models/filters and then other candidate DAGs over the canary input’s frames and stores these results in a table. During query optimization, \tool queries the table only with each DAG’s predicates to produce a final set of labels. The results from the user’s initial DAG
are used as the ground-truth labels. Finally, the candidate
DAG’s F1 score is computed by comparing these labels. 

We use a standard approach of estimating costs: we compute the
latency of executing each DAG for a batch of input frames. The costs of different DAGs are compared to find the most efficient candidate (that still meets the accuracy).

\subsection{Extension}
\label{sec:backend-ext}

We build our backend as an extensible optimization engine for easy integration of emerging video analytics optimizations. Basing \tool off Python enables developers to easily integrate customized optimizations with Python annotations that can meet their desired performance-accuracy tradeoff. Here we showcase how to integrate three optimizations used commonly in previous works: (1)  specialized NNs, (2) binary classifiers for objects, and (3) frame filters.

\MyPara{Specialized NNs.}
Specialized NNs are dedicated to the detection of specific objects and are often much less compute-intensive than general object detectors. Using specialized NNs first in the pipeline to filter out irrelevant frames is an effective optimization, which has been widely adopted in previous systems~\cite{abae, noscope, probpredicate, viva, focus, tahoma}. 

\begin{figure} [ht!]
\includegraphics[width=\linewidth]{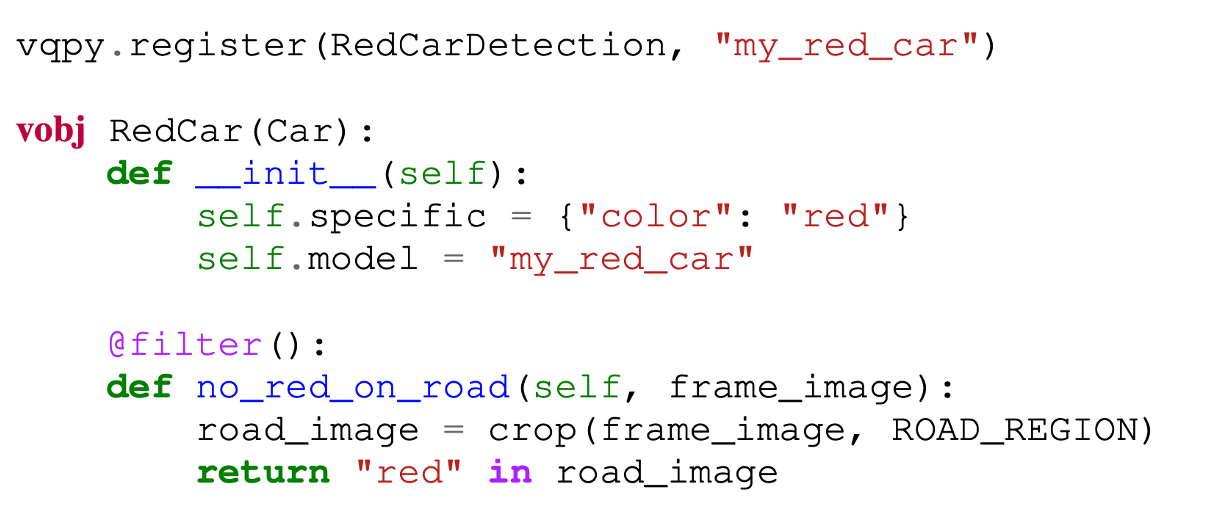}
\vspace{-1em}
\caption{Optionally register models/filters on a \codeIn{RedCar} \vobj.\label{fig:ext-vobj}}
\vspace{-1em}
\end{figure}

Figure~\ref{fig:ext-vobj} shows how to register a specialized NN for the \codeIn{RedCar} \vobj. Users need to first register their specialized NN (\codeIn{RedCarDetection}) into \tool's library by invoking the \codeIn{register} function, so that they can use it in the \vobj by referring to its name (``my\_red\_car''). Models from popular model zoos such as Huggingface, TorchVision, MMLab, and \etc are natively supported in \tool and can be directly registered as specialized NNs.

Register specialized NNs in the super-\vobj can lead to multiple query plans, thus providing additional optimization opportunities. For example, to detect a red car, we could directly employ the specialized red car detector registered with the red car \vobj, or use the general car detector registered with its parent \vobj car and then apply a red color filter. The planner can determine whether to use the specialized model by examining metrics such as the confidence scores of the output results and the cost of different execution paths.

\MyPara{Binary classifiers.}
Binary classifiers directly answer whether an object exists on a frame. With binary classifiers, frames with a low probability of including the target objects are discarded, improving computation efficiency. Binary classifiers have been used in multiple video optimization systems ~\cite{reducto,probpredicate,core}.

Figure~\ref{fig:ext-vobj} also shows how to register a binary classifier (\codeIn{no\_red\_on\_road}) on the \codeIn{RedCar} \vobj, with a simple annotation \codeIn{filter}. \tool's planner takes this information to generate a corresponding filter operator and inserts it at the beginning of the pipeline. The filter inserted calls the \codeIn{no\_red\_on\_road} function to discard frames without any red cars at an early stage, improving computation efficiency. 

\MyPara{Frame Filters.}
Differencing-based frame filters are an effective optimization adopted by many systems~\cite{reducto, noscope,blazeit}; it filters out less informative frames that are close to the background or other frames.

\begin{figure} [htbp]
\includegraphics[width=\linewidth]{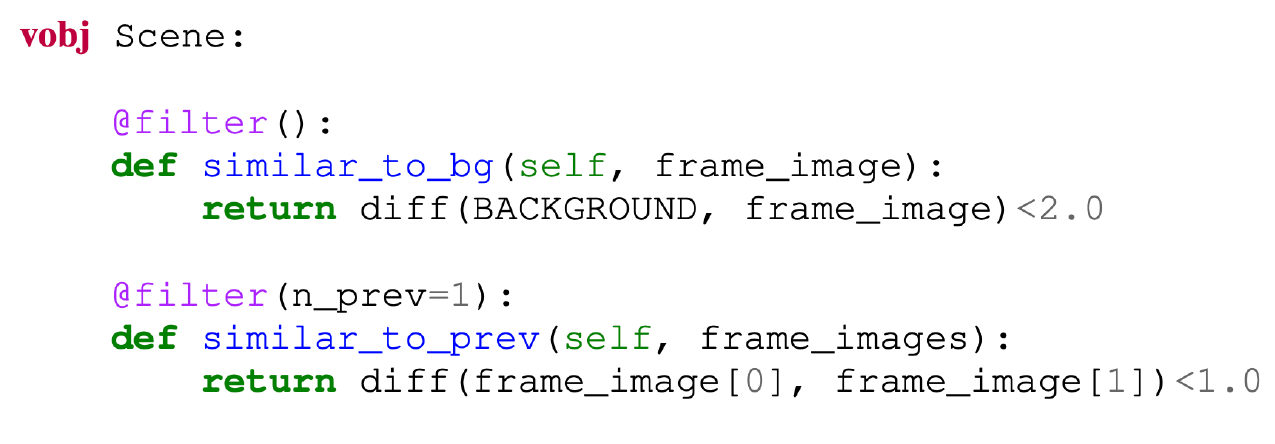}
\vspace{-1em}
\caption{Optionally register frame filters on the \codeIn{Scene} \vobj.\label{fig:ext-frame}}
\vspace{-1em}
\end{figure}

Figure~\ref{fig:ext-frame} shows an example for registering differencing-based frame filters to \tool, which can be defined on \tool's special \codeIn{Scene} \vobj.
To define the \codeIn{similar\_to\_prev} frame filter, which requires the results of a number of previous frames to compare against, users can specify such a number in the \codeIn{filter} annotation. 

\section{Evaluation}
\label{sec:evaluation}

We evaluated \tool on 14 queries on 5 datasets from real-world surveillance video streams. Our evaluation results demonstrate that:

(1) \tool achieves up to 12.6$\times$ query speedups compared to manually crafted pipelines on 5 complex vehicle retrieval queries (\S\ref{sec:eval-city}). 

(2) \tool achieves up to 15.2$\times$ speedups compared to EVA, the state-of-the-art SQL-based framework, on video object queries (\S\ref{sec:eval-sql}). 

(3) Compared to a MLLM-based approach VideoChat, \tool is 15.18$\times$ faster, requires 90$\%$ percent less
GPU memory usage, and produces significantly (3-5$\times$) higher accuracy (\S\ref{sec:eval-mllm}). 

\S\ref{sec:eval-adopt} briefly discusses how \tool is adopted by a major tech company to help its customers develop complex queries.

\subsection{Comparison with Handcrafted Pipelines
\label{sec:eval-city}}

\MyPara{Dataset and Queries.}
Our experiments utilized the CityFlow-NL dataset~\cite{cityflow-nl} from the 2023 AI CITY CHALLENGE, Challenge Track 2~\cite{Naphade23AIC23}, featuring 3.25 hours of traffic footage across 10 intersections from 40 cameras, with a minimum resolution of 960p at 10 frames per second. We evaluated on all 36 videos in the test set, which includes 184 vehicle tracks paired with natural language queries describing various vehicle attributes and scenarios. 
The original challenge was framed as a vehicle retrieval problem where vehicles for these queries are ranked. We repurposed the task into a video analytics problem which is concerned with the locations of the video frames that include the vehicles meeting the query constraints. Specifically, we randomly chose 5 queries from the CityFlow-NL dataset listed in Table~\ref{tab:city-query}.

\begin{table}[t]\centering
\scriptsize
\begin{tabular}{p{0.1in}p{1.6in}p{1in}}\toprule

\textbf{No.} &\textbf{NL Descriptions} &\textbf{CVIP standardized} \\\midrule
\textbf{Q1} & "A green sedan is keeping straight." &"green sedan go straight" \\
\textbf{Q2} & "A green bus going straight down the street followed by a white car." &"green bus go straight" \\
\textbf{Q3} &"A red sedan runs down the street." &"red sedan go straight" \\
\textbf{Q4} & "A black sedan keeps driving forward." &"black sedan go straight" \\
\textbf{Q5} & "A large black SUV turns right." &"black suv turn right" \\
\bottomrule
\end{tabular}
\vspace{-1em}
\caption{Queries selected from CityFlow-NL.\label{tab:city-query}}
\vspace{-2em}
\end{table}

\MyPara{Settings.}
We compared our method with CVIP~\cite{Le_2023_CVPR}, the top prize winner in this Challenge Track. As detailed in Table~\ref{tab:city-query}, CVIP standardizes the natural language queries into a fixed format of \emph{color-type-direction} during preprocessing. We adapted \tool to use the same standlized query format, and evaluated the runtime performance of both \tool and CVIP excluding the text standardization pre-processing step.
Note that \tool focuses on query development and optimization, \emph{not} on model improvement (\ie, for higher accuracy). 
For a fair comparison, we let \tool use the same pre-trained vision models as used by CVIP in each query.
We used two configurations for \tool: the vanilla \tool and \tool with user-provided intrinsic annotations (see \S\ref{sec:backend-reuse}) for \codeIn{color} and \codeIn{type}.
We evaluated both \tool and CVIP on a Google Cloud virtual machine instance, equipped with one NVIDIA T4-16G GPU, 16 vCPUs (8 cores), and 104GB RAM.

\begin{figure}[!t]
	\centering
        \subfigcapskip=-1.5ex
        \hspace{-                                                               2ex}
	\subfigure[Runtime comparison.]{
		\includegraphics[width=0.5\linewidth]{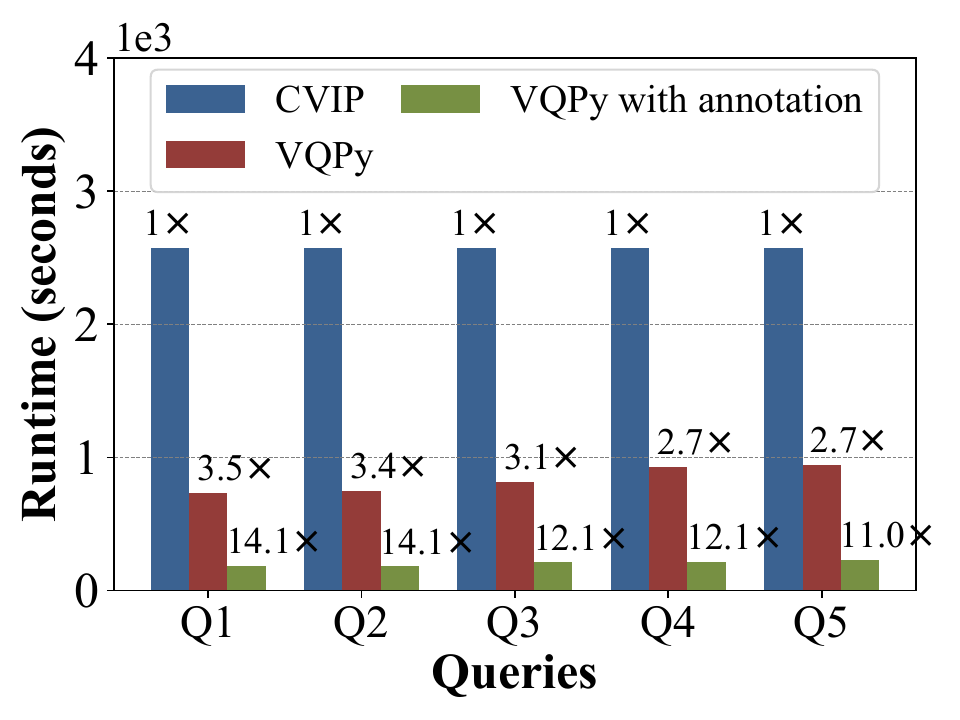}
		\label{fig:compared_with_CVIP}
	}
        \hspace{-1.5ex}
	\hfil
	\hspace{-1.5ex}
	\subfigure[Time spent on each frame.]{
		\includegraphics[width=0.5\linewidth]{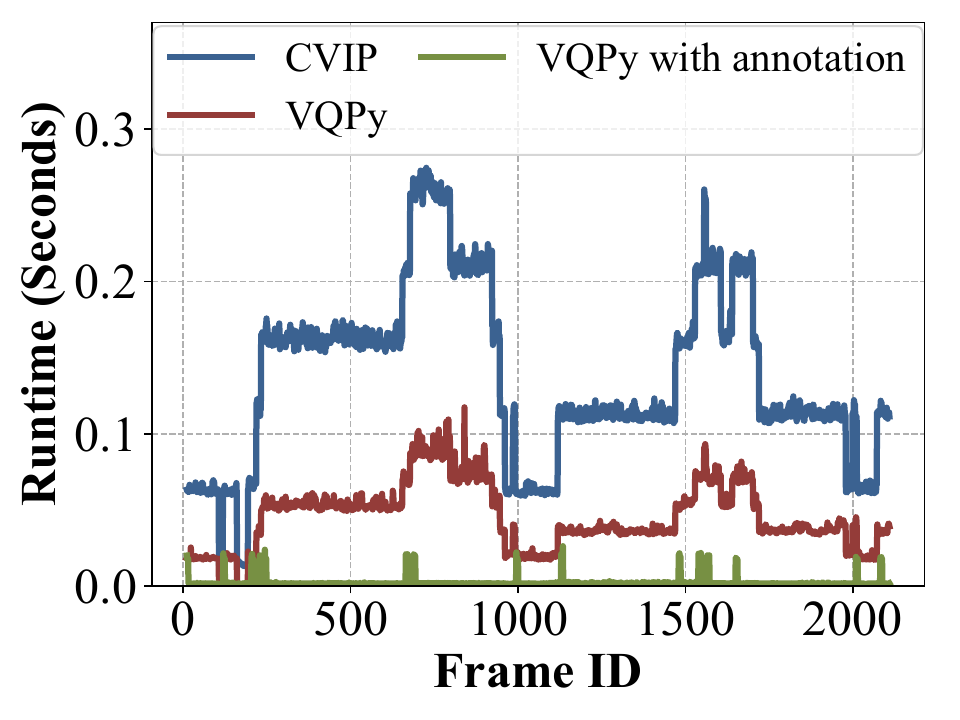}
		\label{fig:compared_with_CVIP2}
	}
        \hspace{-2ex}
	\vspace{-2ex}
	\caption{Performance Comparison of \tool and CVIP.}
	\vspace{-5ex}
	\label{fig:compared_with_CVIP_12}
\end{figure}

\MyPara{Results.}
\tool achieves the same accuracy as CVIP across all five queries, due to the use of the same pretrained models. Figure~\ref{fig:compared_with_CVIP} compares the query execution time between \tool and CVIP.
Regardless of the query type, CVIP consistently requires around 850 seconds due to the necessity of processing \emph{all} cropped images with all detection models (for color, type, and direction), resulting in a stable runtime.
In contrast, VQPy employs filters after the computation of each property, efficiently filtering out vehicles that do not meet a property condition before computing other properties. For example, if the \codeIn{color} property of a frame does not satisfy the condition (\eg, red), \tool will not proceed to the computation of the other properties. 
This approach significantly reduces unnecessary computations, resulting in an average \textbf{3.1$\times$} speedup over CVIP, as shown in Figure~\ref{fig:compared_with_CVIP}. This performance gain stems from VQPy's object-centric approach (\eg, properties can be computed and associated with individual video objects) as well as its use of \emph{lazy evaluation}.

Specifically, \tool demonstrates more pronounced performance gains for queries on green vehicles than black vehicles, as the former are less common in the dataset. This rarity results in more objects being filtered out early in the detection process, reducing unnecessary model inferences.
Moreover, VQPy with user-specified intrinsic annotations outperforms vanilla VQPy by allowing reuse of computation results for \codeIn{color} and \codeIn{type}.

The detailed savings in per-frame computation time are reported in Figure~\ref{fig:compared_with_CVIP2}, which depicts how the per-frame computation changes as frames are being processed. As seen, though spending much less time on each frame than CVIP (due to the use of lazy evaluation), the vanilla \tool follows a similar computation curve while the intrinsic annotations flatten the curve, due to computation reuse. 
Overall, intrinsic annotations provide an additional \textbf{9.5$\times$} speedup compared to the vanilla \tool, bringing the improvement over CVIP up to \textbf{12.6$\times$}.

\subsection{Comparisons with SQL-based Frameworks}
\label{sec:eval-sql}

\MyPara{Baselines.}
We compared \tool with the state-of-the-art SQL-based framework,  EVA~\cite{eva}, whose optimizations subsume those in other frameworks such as \cite{blazeit, noscope, probpredicate}. EVA is also the most well-developed framework, supporting the largest number of SQL primitives.

\MyPara{Queries.}
It is hard to construct video queries involving complex relationships between objects with EVA. Therefore we compared \tool and EVA only with queries that count video objects with specific properties. We chose three types of queries: objects with stateless properties, objects with stateful properties, and objects with both kinds of properties.

\begin{table}[th]\centering
\scriptsize
\begin{tabular}{lllrr}\toprule
\textbf{Query type} &\textbf{Specific Query} &\textbf{Eva \& VQPy expression} \\\midrule
Stateless property &Red car & See Figures \ref{fig:eva-red-car}, \ref{fig:vqpy-red-car}  \\
Stateful property &Speeding car & See Figures \ref{fig:eva-speed-car}, \ref{fig:vqpy-speed-car} \\
Stateless \& stateful properties & Red speeding car & See Figures \ref{fig:eva-red-speed-car}, \ref{fig:vqpy-red-speed-car} \\
\bottomrule
\end{tabular}
\vspace{-1em}
\caption{Stateful and stateless properties to compare with EVA.}\label{tab:sql-query}
\vspace{-1em}
\end{table}

Table~\ref{tab:sql-query} summarizes the stateful and stateless properties we chose. For a fair comparison, we let \tool use EVA's built-in YOLO detection model and nor-fair tracker. To detect color, we adopted an NN model from CVIP. The model can be easily annotated as a stateless property of color on the car \vobj. To integrate the same model into EVA, we wrote a UDF to wrap the model around adapting the I/O (pandas Dataframes) formats required by EVA. To detect speed, we handcrafted a function using information from objects' bounding boxes. This function was used directly as a stateful property in \tool; we turned it into a Python UDF to integrate it in EVA. 

EVA lacks support for important SQL primitives such as window functions or group-by, which are necessary for forming the input, often a set of objects in consecutive frames, for stateful properties of video objects (\eg, the speed of cars). 
To retrieve this historical data for each video object in Eva, we had to join the tables from different (\textit{i-th}) lagged frames.

Appendix~\S\ref{sec:appendix-A} shows the actual query code under \tool and EVA: it is clear that \tool enables straightforward development of video queries while EVA presents users a relational structure requiring much more mental power to translate video objects and tables back and forth.  

\begin{table}[th]\centering
\scriptsize
\begin{tabular}{lrrrr}\toprule
\textbf{Camera location} &\textbf{FPS} &\textbf{Resolution} \\\midrule
Banff, Candada~\cite{banff} &15 &1280x720  \\
Jaskson Hole, WY~\cite{jackson} &15 &1920x1080  \\
Southampton, NY~\cite{southampton} &30 &1920x1080 \\
\bottomrule
\end{tabular}
\vspace{-1em}
\caption{Video datasets to compare with SQL-based frameworks.}\label{tab:sql-datasets}
\vspace{-1em}
\end{table}

\MyPara{Datasets.}
Table ~\ref{tab:sql-datasets} summarizes our datasets, which come from public surveillance video streams deployed around North America and have been adopted by other works~\cite{reducto, eva, blazeit, noscope} in evaluation. 
To evaluate how different video lengths affect the query execution time, we constructed the video datasets with two configurations: (1) 5 video clips of 3 minutes each, and (2) 5 video clips of 10 minutes each. 
Our evaluations were conducted on a Google Cloud
virtual machine instance with an NVIDIA T4-
16G GPU, 16 vCPUs (8 cores), and 104GB RAM.

\begin{figure}[!t]
	\centering
        \subfigcapskip=-1.5ex
        \hspace{-2ex}
	\subfigure[3 min.]{
		\includegraphics[width=0.5\linewidth]{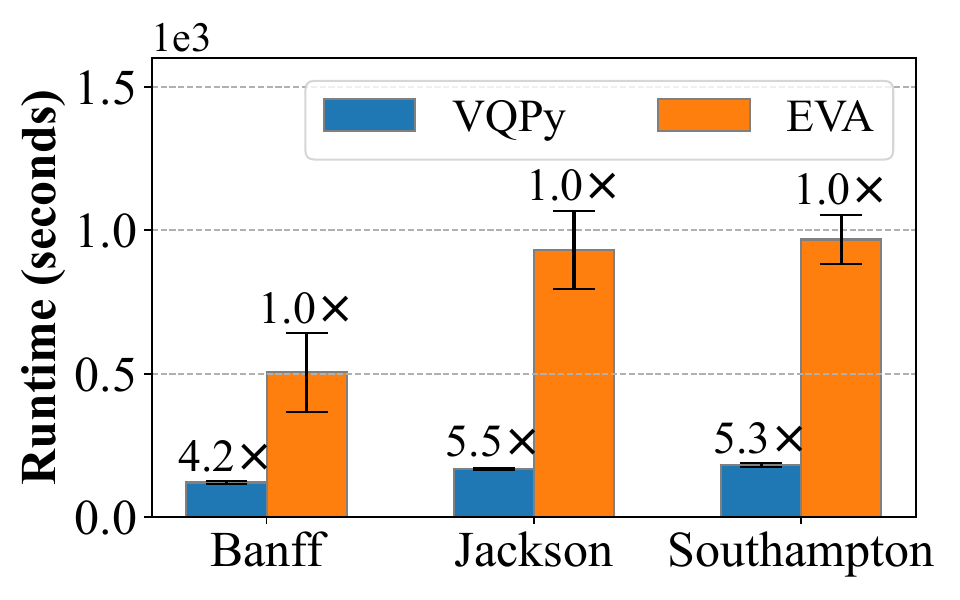}
		\label{Q1_3min}
	}
        \hspace{-2ex}
	\hfil
	\hspace{-2ex}
	\subfigure[10 min.]{
		\includegraphics[width=0.5\linewidth]{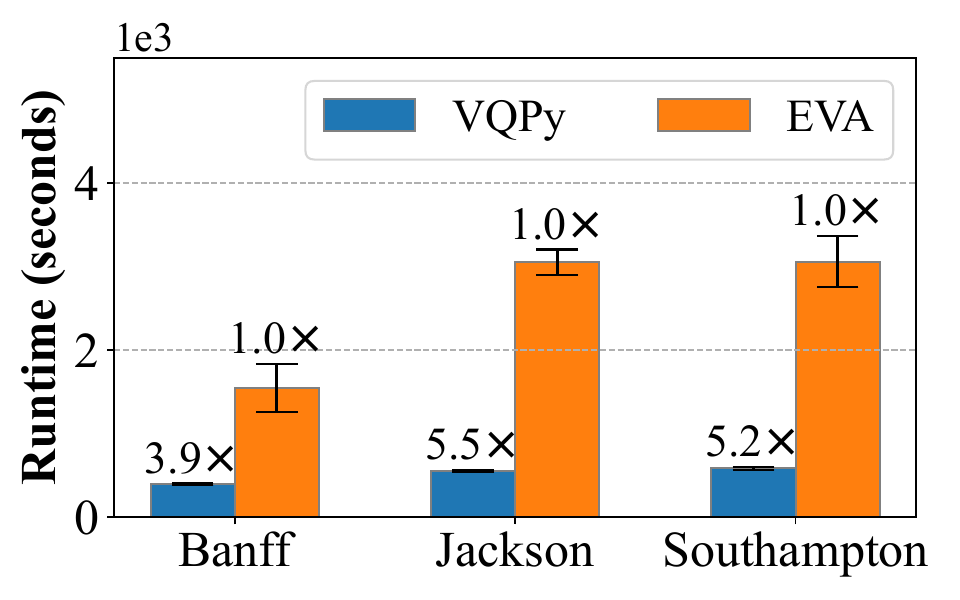}
		\label{Q1_10min}
	}
        \hspace{-2ex}
	\vspace{-2.5ex}
	\caption{Red Car Query: VQPy is averagely 4.9× faster.}
	\vspace{-4ex}
	\label{fig:vqpy-red-car-result}
\end{figure}

\begin{figure}[!t]
	\centering
        \subfigcapskip=-1.5ex
        \hspace{-2ex}
	\subfigure[3 min.]{
		\includegraphics[width=0.5\linewidth]{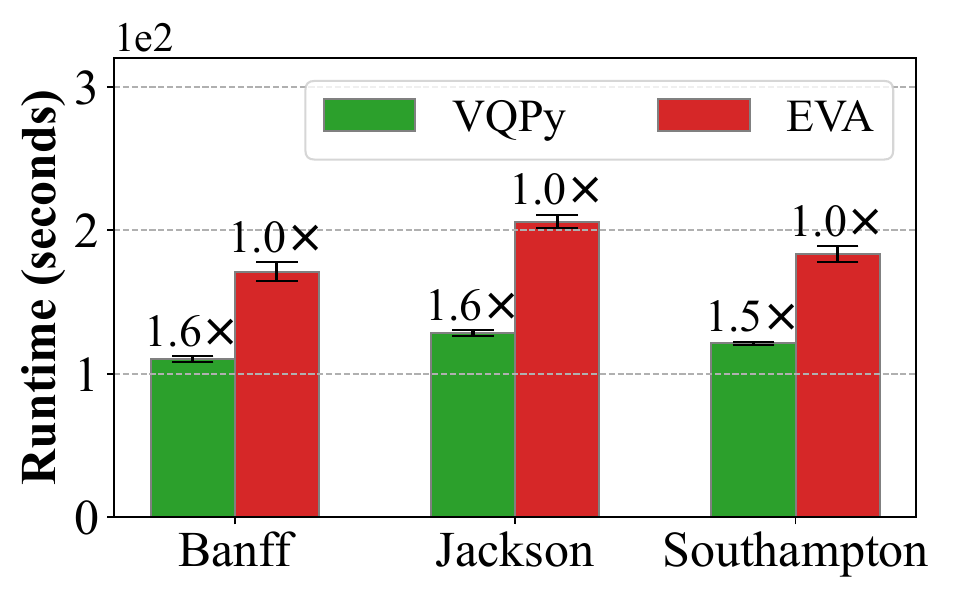}
		\label{Q2_3min}
	}
        \hspace{-2ex}
	\hfill
	\hspace{-2ex}
	\subfigure[10 min.]{
		\includegraphics[width=0.5\linewidth]{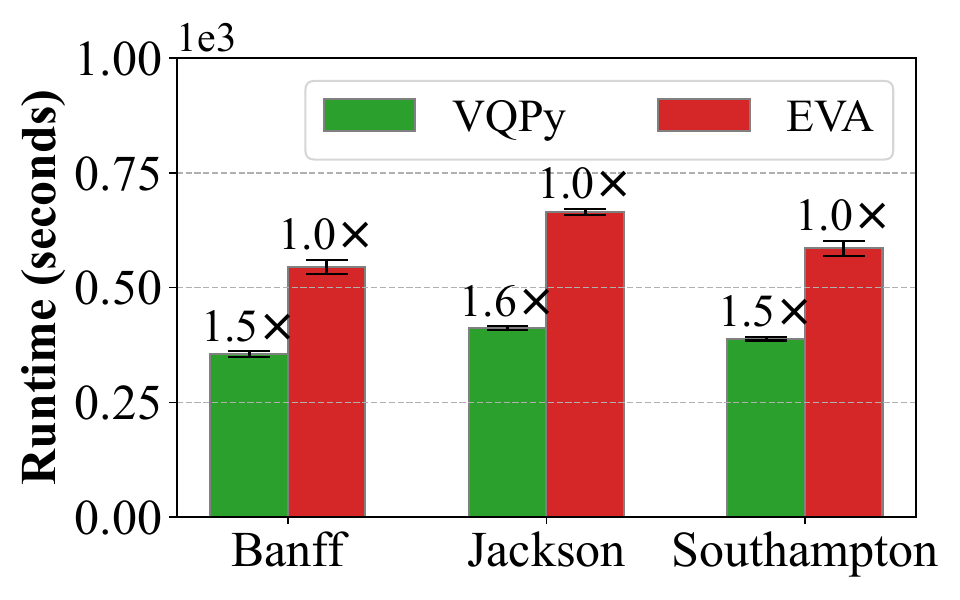}
		\label{Q2_10min}
	}
        \hspace{-2ex}
	\vspace{-2.5ex}
	\caption{Speeding Car Query: VQPy is averagely 1.5× faster.}
	\vspace{-4ex}
	\label{fig:vqpy-speed-car-result}
\end{figure}

\begin{figure}[!t]
	\centering
        \subfigcapskip=-1.5ex
        \hspace{-2ex}
	\subfigure[3 min.]{
		\includegraphics[width=0.5\linewidth]{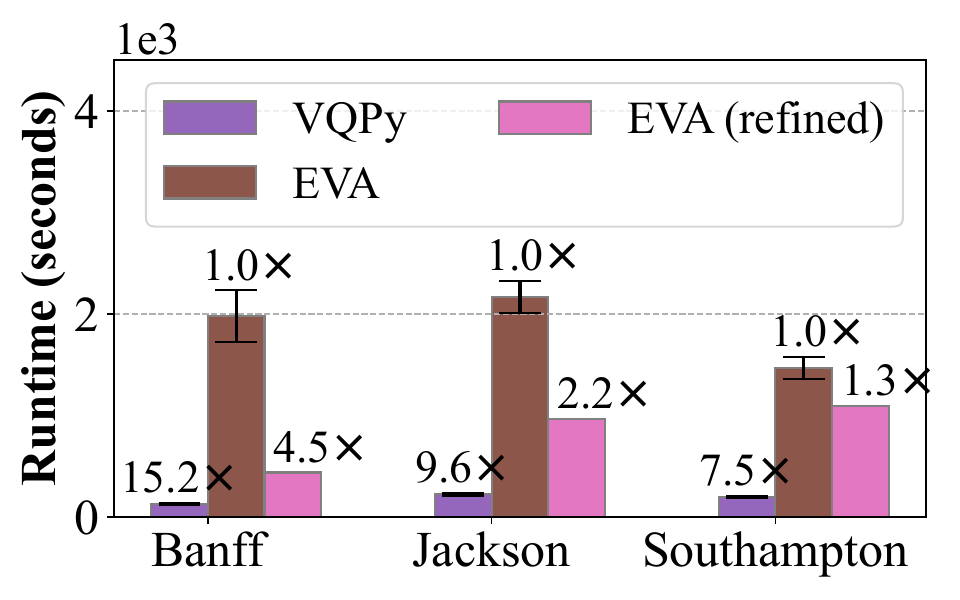}
		\label{Q3_3min}
	}
        \hspace{-2ex}
	\hfill
	\hspace{-2ex}
	\subfigure[10 min.]{
		\includegraphics[width=0.5\linewidth]{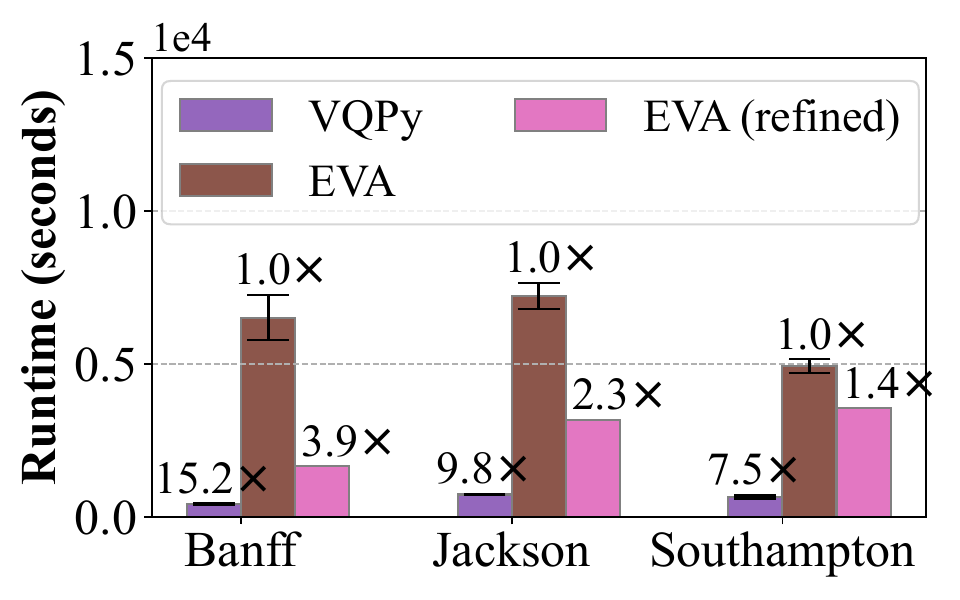}
		\label{Q3_10min}
	}
        \hspace{-2ex}
	\vspace{-2.5ex}
	\caption{Red Speeding Car Query: VQPy is averagely 11× faster.}
	\vspace{-4ex}
	\label{fig:vqpy-red-speed-car-result}
\end{figure}

\MyPara{Results.}
Since EVA does not contain any frame filters or specialized NNs, we also disabled such optimizations in \tool for a fair comparison. As such, query accuracy depends purely on the model chosen and the UDFs. By using the same models and UDFs, EVA and \tool achieve the same accuracy. 
The performance comparisons are reported in Figure~\ref{fig:vqpy-red-car-result}, Figure~\ref{fig:vqpy-speed-car-result} and Figure~\ref{fig:vqpy-red-speed-car-result}, respectively, for the three queries. Each figure consists of the results for both 
3-minute video clips (left) and 10-minute clips (right). 

\textit{Stateless property.} As shown in Figures~\ref{Q1_3min} and \ref{Q1_10min}, \tool achieves an average of \textbf{{5.0}$\times$} speedup on the short video clips and a \textbf{{4.8}$\times$} speed up on the long clips. This is primarily due to the computation reuse across the frames for the same \vobj, for its intrinsic \codeIn{color} property.

\textit{Stateful property.}  As shown in Figure~\ref{fig:vqpy-speed-car-result}, \tool is \textbf{{1.5}$\times$} faster than EVA where the performance gain comes from EVA's requirement of using expensive table joins to compute the stateful property of \codeIn{speed}. 

\textit{Stateless and stateful property.} 
EVA allows each query to contain only one single statement. As such, to implement queries on video objects with both stateless and stateful queries, we had to use query nesting. As shown in Figure~\ref{fig:vqpy-red-speed-car-result}, EVA is \textbf{7.5}-\textbf{15.2$\times$} slower compared to VQPy. In addition to the aforementioned reasons, EVA does not support creating ``VIEW'' from queries, and hence filters used in later part of the query cannot be pushed to apply on earlier tables, leading to redundant executions of UDFs. 
We manually optimized EVA's SQL queries by pushing down the filters. Despite this optimization, EVA is still \textbf{3.3-5.7$\times$} slower, due to its inability to perform object-level optimizations. 
\subsection{Comparisons with Multimodal LLMs}
\label{sec:eval-mllm}

\begin{table}[!htp]\centering
\scriptsize
\begin{tabular}{p{0.1in}p{0.5in}p{2.15in}}\toprule
\textbf{No.} &\textbf{Query Type} &\textbf{Statement} \\\midrule
\multicolumn{2}{l}{\textbf{Prompt}} & You are an AI assistant. A human gives a video about traffic and asks questions about the humans and cars on it. You should give helpful, detailed, and polite answers. \\
\textbf{Q1} &Boolean & Are there any people passing the crosswalk? \\
\textbf{Q2} &Boolean & Are there any cars turning left at the crossing? \\
\textbf{Q3} &Boolean & Are there any red cars in the video? \\
\textbf{Q4} &Aggregation & Tell me the average number of cars on the crossing. \\
\textbf{Q5} &Aggregation & Tell me the average number of people that are walking. \\
\textbf{Q6} &Boolean & Is anyone hitting the ball in the image? Answer by yes or no. \\
\bottomrule
\end{tabular}
\vspace{-1em}
\caption{Three query sets (Q1-Q3, Q4-Q5, Q6) used to compare \tool with VideoChat, and their natural language statement.\label{tab:mllm-query}}
\vspace{-2em}
\end{table}

\vspace{0.6em}

\begin{figure}[htbp]
\centering
\includegraphics[width=0.4 \textwidth]{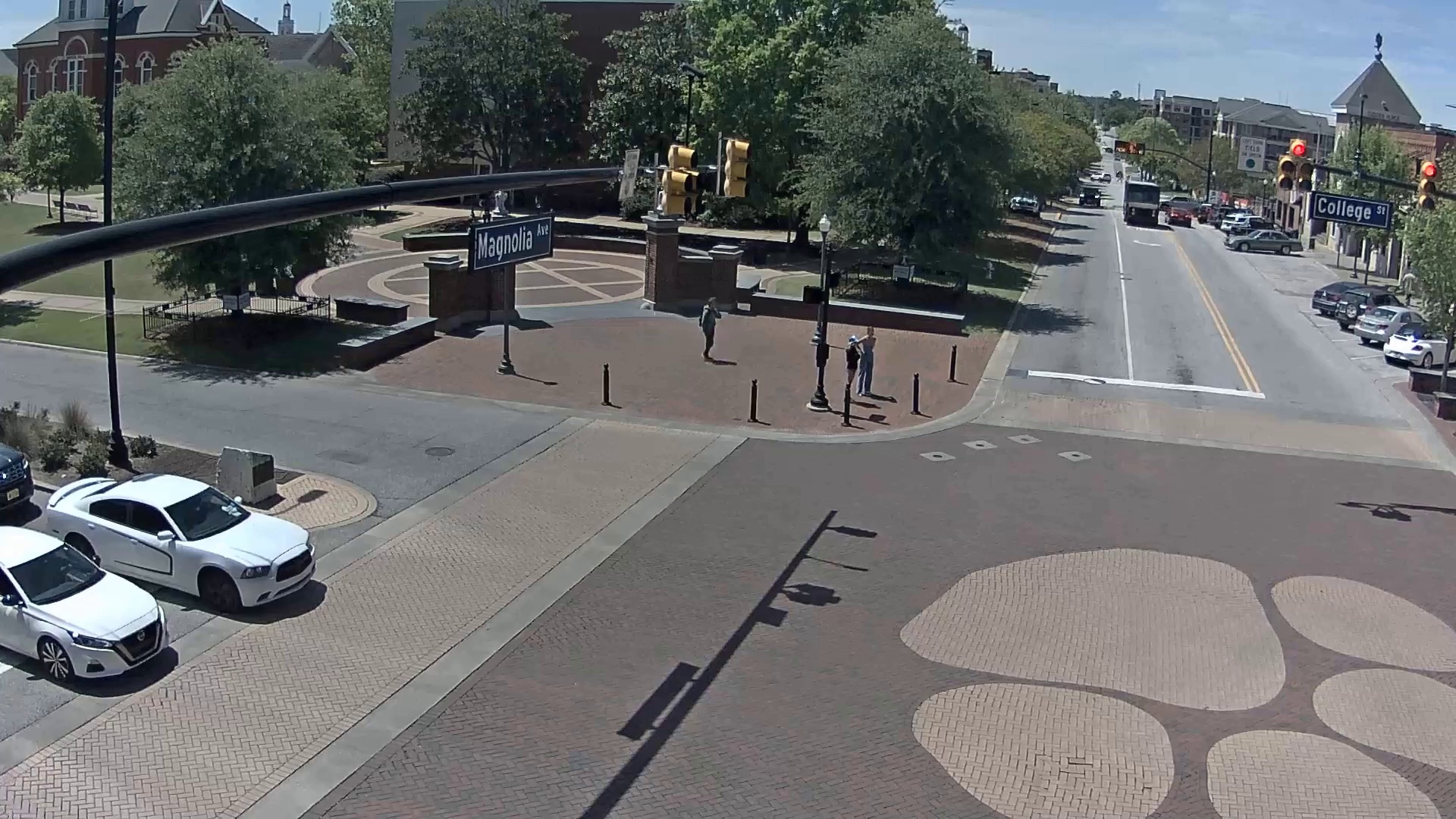}
\vspace{-1em}
\caption{Sample image from Auburn video. \label{fig:auburn-frame}}
\vspace{-1.5em}

\end{figure}

\MyPara{Queries and Datasets.} We have constructed a set of 6 queries to compare \tool with MLLM-based methods, summarized in Table~\ref{tab:mllm-query}. This set includes queries on specific video objects, aggregation queries, and those involving object interactions. We used Auburn \cite{auburndata}, a public surveillance video that monitors the traffic at a crossroad, as our dataset for the first two queries. Table \ref{tab:sql-datasets} includes some of its properties, and a sample frame from the video can be found in Figure \ref{fig:auburn-frame}. Since MLLM models consume a large amount of computation resources, we could only use a 10-minute clip in the daytime in the experiment. The first query set (Q1-Q3) contains three boolean queries asking for Yes/No responses. The second query set (Q4 and Q5) consists of aggregation queries asking for a floating value.

For Q6, we used an image dataset V-COCO \cite{vcoco} to query on object relationships. This dataset, which contains 4532 images for testing, selects a subset of MS-COCO \cite{coco} images and annotates the human-object interaction. In our evaluation, we considered only one specific human-object pair and queried whether this interaction exists in each clip.

\begin{figure}[htbp]
\centering
\includegraphics[width=0.48 \textwidth]{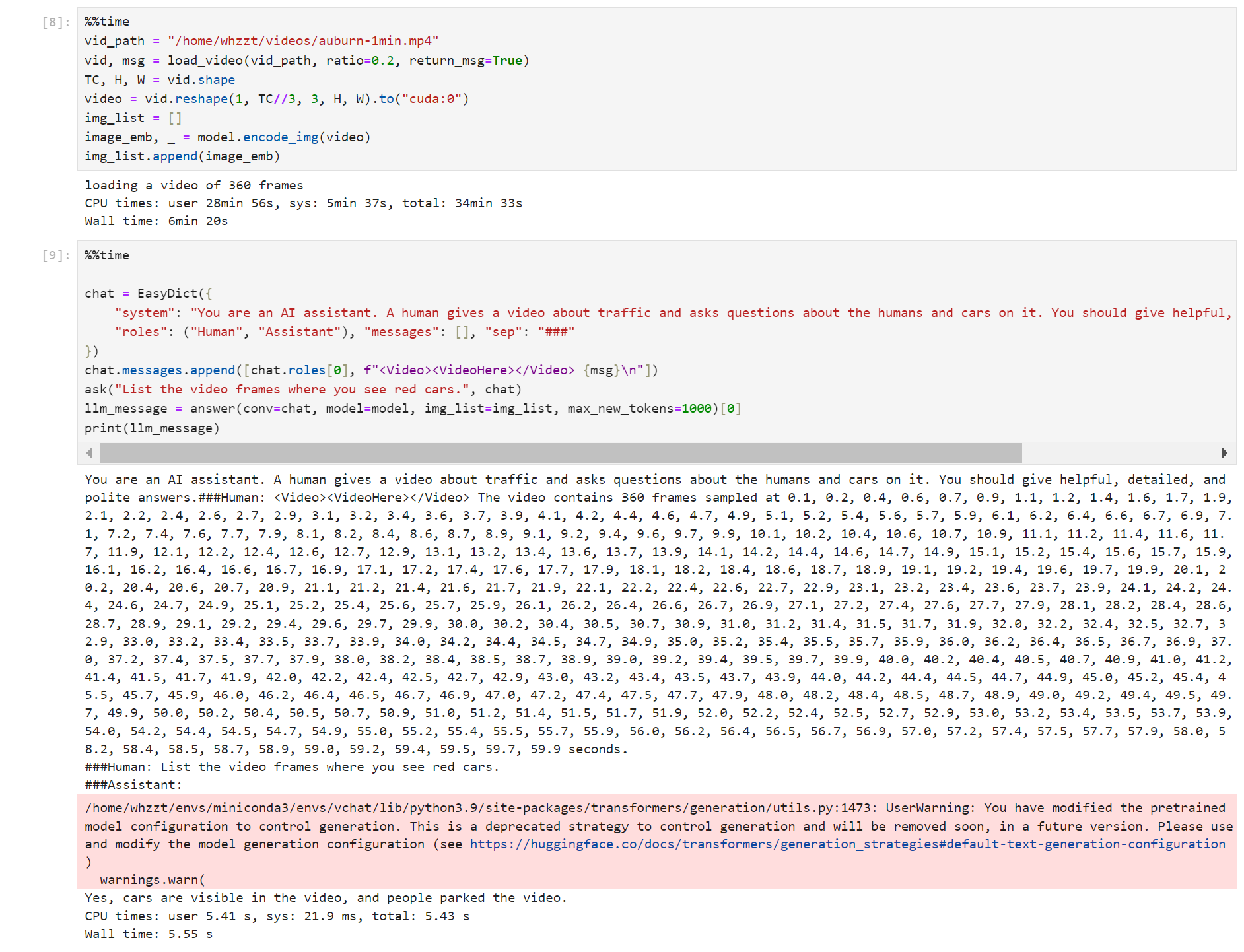}
\vspace{-1em}
\caption{VideoChat is not able to solve long video on-frame queries. \label{fig:mllm-failure}}
\vspace{-1.5em}
\end{figure}

\MyPara{VideoChat.}
We compared \tool with VideoChat~\cite{li2023videochat}, a state-of-the-art multimodal LLM designed for video analytics, which can support both boolean queries and aggregation queries. However, it can only answer questions regarding the entire video, \emph{not} individual frames. When asking questions regarding individual frames, VideoChat often provided irrelevant responses; an example of that is illustrated in Figure~\ref{fig:mllm-failure}. When the length of the video grows, the consumption of GPU memory rapidly increases: we need at least 40GB GPU memory to process VideoChat-7b for a video clip as short as 540 frames of a 1920x1080 resolution. Moreover, the video embedding computation is slow, and this problem becomes more pronounced when we offload parts of the model to the CPU to save GPU memory.

Consequently, we had to split the 10-minute video dataset into 600 one-second clips and query VideoChat on each clip. 

\MyPara{Evaluation Setting.} We evaluated both \tool and VideoChat on a Google Cloud virtual machine instance, equipped with one A100-40G GPU, 12 vCPUs (6 cores), and 85GB RAM. We considered two versions of VideoChat in our experiments: VideoChat-13B and VideoChat-7B. Since our GPU memory is insufficient to store both the entire VideoChat-13B model and the intermediate results, we enabled the low-resource mode when running the VideoChat-13B model, which uses 8-bit weight and offloads part of the video embedding computation to CPU memory.

Since VideoChat can only provide natural language replies, we had to analyze the query results ourselves. We carefully designed the clear statements (as shown in Table~\ref{tab:mllm-query} to make the responses more regular. We used a pattern-based analyzer to resolve most of the responses and annotated the remaining manually. For unclear responses, we simply dropped these data points when computing accuracy.

For the first two sets of queries Q1-Q5, we used YOLOX \cite{ge2021yolox} as \tool's detection model. For the interaction query Q6, we used UPT \cite{upt}, one of the best two-stage models in the V-COCO dataset as \tool's baseline model.

\begin{table}[htbp]\centering
\scriptsize
\begin{tabular}{p{0.1in}p{0.7in}p{0.7in}p{0.4in}p{0.5in}}\toprule
\textbf{No.} &\textbf{VideoChat-7B} &\textbf{\textbf{VideoChat-13B$^{*}$}} & \textbf{\tool} & \textbf{\tool-Opt} \\\midrule
\textbf{Pre} & 38.4 & 1071.0 & N/A & N/A \\
\midrule
\textbf{Q1} & 72.4 & 656.3 & \textbf{34.4} &  \\
\textbf{Q2} & 80.7 & 637.3 & \textbf{32.9} &  \\
\textbf{Q3} & 85.1 & 563.7 & \textbf{48.2} & \textbf{52.6} \\
\textbf{Q4} & 116.9 & 848.6 & \textbf{31.9} &  \\
\textbf{Q5} & 137.3 & 836.8 & \textbf{35.4} &  \\
\midrule
\textbf{Q6} & 3503.8 & 8183.5 & 112.4 & \textbf{30.0}\\
\bottomrule
\end{tabular}
\vspace{-1em}
\caption{Execution time for VideoChat and \tool (millisecond per frame). Note that VideoChat has a pre-computation phase that loads the video and computes its embedding. \tool-Opt combines Q1-Q6 in a single execution with computation reuse enabled. \label{tab:mllm-performance} \\
VideoChat-13B$^{*}$: Low resource mode due to GPU memory limit.}
\vspace{-2em}
\end{table}

\vspace{1em}

\begin{table}[htbp]\centering
\scriptsize
\begin{tabular}{p{0.1in}p{0.5in}p{0.7in}p{0.7in}p{0.4in}}\toprule
\textbf{No.} & \textbf{Pr(positive)} &\textbf{VideoChat-7B} &\textbf{VideoChat-13B$^{*}$} & \textbf{VQPy} \\\midrule
\textbf{Q1} & 21.7\% & 0.412 & 0.422 & \textbf{0.902} \\
\textbf{Q2} & 37.5\% & 0.382 & 0.360 & \textbf{0.591} \\
\textbf{Q3} & 46.1\%  & 0.674 & 0.685 & \textbf{0.915} \\
\textbf{Q6} & 4.9\% & 0.130 & 0.237 & \textbf{0.867} \\
\bottomrule
\end{tabular}
\vspace{-1em}
\caption{F-1 score for boolean queries. We also provide the positive sample rate in the dataset to reflect the structure of data. \label{tab:mllm-accuracy-boolean}\\
VideoChat-13B$^{*}$: Low resource mode due to GPU memory limit.}
\vspace{-1em}
\end{table}

\MyPara{Results.} Table~\ref{tab:mllm-performance} reports the execution time for VideoChat and \tool under these queries. \tool outperforms VideoChat under all the settings and tests.

Table~\ref{tab:mllm-accuracy-boolean} shows the F-1 score for the boolean queries Q1, Q2, Q3 and Q6. For Q1-Q3, we obtained the ground truth by labeling them manually. For Q6, we used the V-COCO annotations to generate the result. We also provided the percentage of queries that have a Yes response, which can affect the F-1 score if the tested program has a fixed probability of misprediction on each type of response.
To summarize, \tool exhibits superior accuracy, achieving an average F-1 score of \textbf{82\%} across queries Q1, Q2, Q3, and Q6, far surpassing the F-1 scores of VideoChat-13B and VideoChat-7B, which are 43\% and 40\%, respectively.

\begin{table}[!htp]\centering
\scriptsize
\begin{tabular}{p{0.7in}p{1in}p{1in}}\toprule
\textbf{Model} & \textbf{Average Response} &\textbf{Maximum Response} \\\midrule
\textbf{VideoChat-7B} & Q4: 6.87; Q5: 6.78 & Q4: 250; Q5: 414 \\
\textbf{VideoChat-13B$^{*}$} & Q4: 4.86; Q5: 4.95 & Q4: 65; Q5: 100 \\
\textbf{VQPy} & Q4: 0.89; Q5: 0.66 & Q4: 3.3; Q5: 5.28 \\
\bottomrule
\end{tabular}
\vspace{-1em}
\caption{Evaluation results for Q4 and Q5. \label{tab:mllm-accuracy-agg}\\
VideoChat-13B$^{*}$: Low resource mode due to GPU memory limit.}
\vspace{-1em}
\end{table}

Table~\ref{tab:mllm-accuracy-agg} summarizes the evaluation results for the aggregation queries Q4 and Q5, revealing that VideoChat's answers are rather inaccurate. For VideoChat-13B, 73.7\% queries are preserved for Q4 and 60.1\% for Q5. For VideoChat-7B, these percentages are 64.2\% and 53.2\%, respectively. For Q4, there are never more than 4 cars on the crossing at the same time, but under both models, the average number reported by VideoChat exceeds this. For Q5, there are never more than 10 walking people in the video, but there are at least 15\% of responses that return a value larger than 10.

\MyPara{Available Optimizations.} Similar to VideoChat which allows users to ask multiple queries after uploading a video, we can also support this in \tool to reduce unnecessary computations. We tested the \tool pipeline that executes Q1-Q5 in a single execution, which results in an overall 3.4$\times$ speedup compared to executing them individually; details are shown in Table~\ref{tab:mllm-performance}.

Another optimization we performed on Q6 is to extend our backend with specialized NNs and filters. We used a cheap detector \cite{yolov5} to filter out frames without target objects, and trained a specialized model (following the ideas in \cite{interactions}) to drop the frames that are unlikely to contain the required action. With these optimizations, we were able to obtain a further gain of \textbf{3.7$\times$} with a 0.08 loss in the F1-score.

\subsection{Real-World Adoption and Use Cases}
\label{sec:eval-adopt}
\tool has been productized in Cisco as a query development/execution layer in its major vision product \emph{DeepVision}, to enable its customers to develop and execute complex queries.
DeepVision is a comprehensive video analytics system that enables users to monitor and analyze video streams from various sources with ease. It is a scalable and modular serverless open-source framework with state-of-the-art object detectors, trackers, and behavior detectors integrated.

\tool was integrated into DeepVision as a service, executing queries over a real-time video stream from the company's video source service. \tool's results are streamed back to DeepVision's dashboard for real-time monitoring. 
\tool empowers DeepVision with the video query ability, orchestrating the underlying models' output, and automatically selecting the most efficient plan to construct and execute the pipeline, satisfying the real-time query latency requirements. 

\begin{figure}[htbp]
     \centering
     \begin{minipage}[t]{0.48\textwidth}
         \centering
         \includegraphics[width=\textwidth]{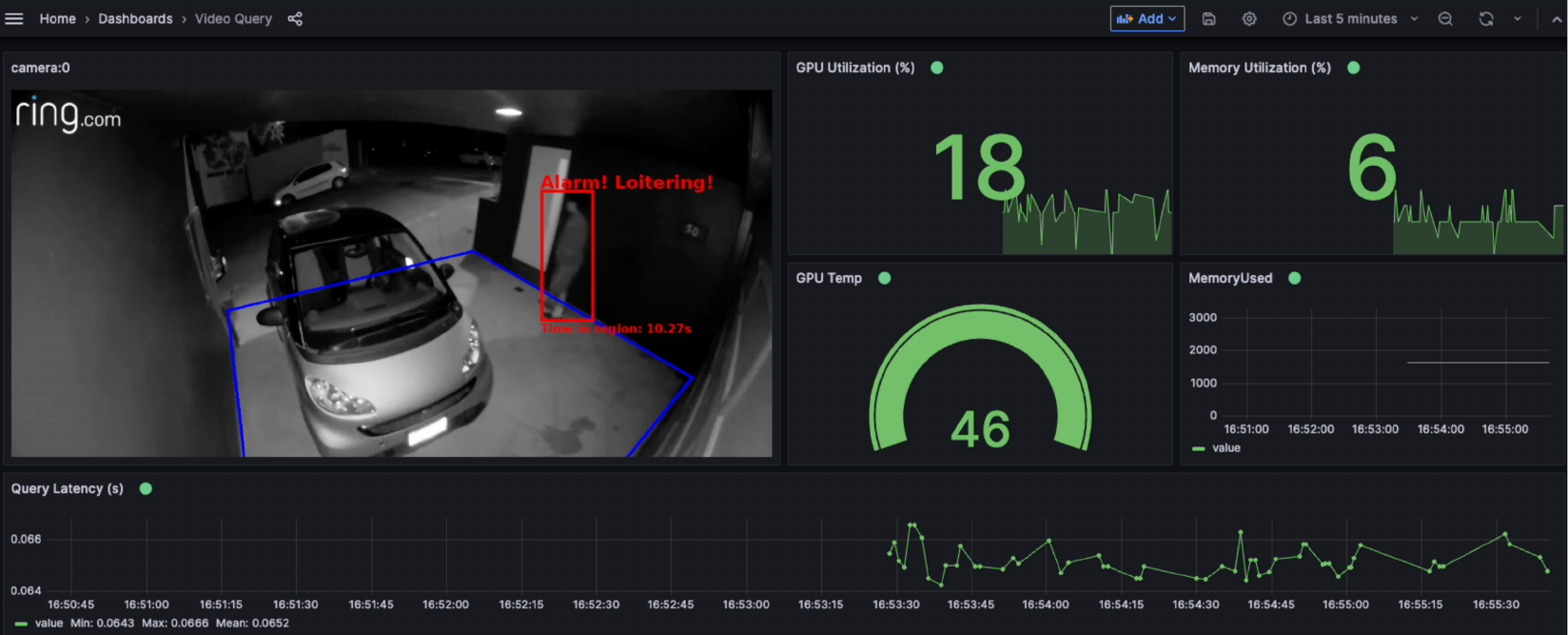} 
         \footnotesize (a) Loitering alert.
     \end{minipage}
     \begin{minipage}[t]{0.48\textwidth}
         \centering
         \includegraphics[width=\textwidth]{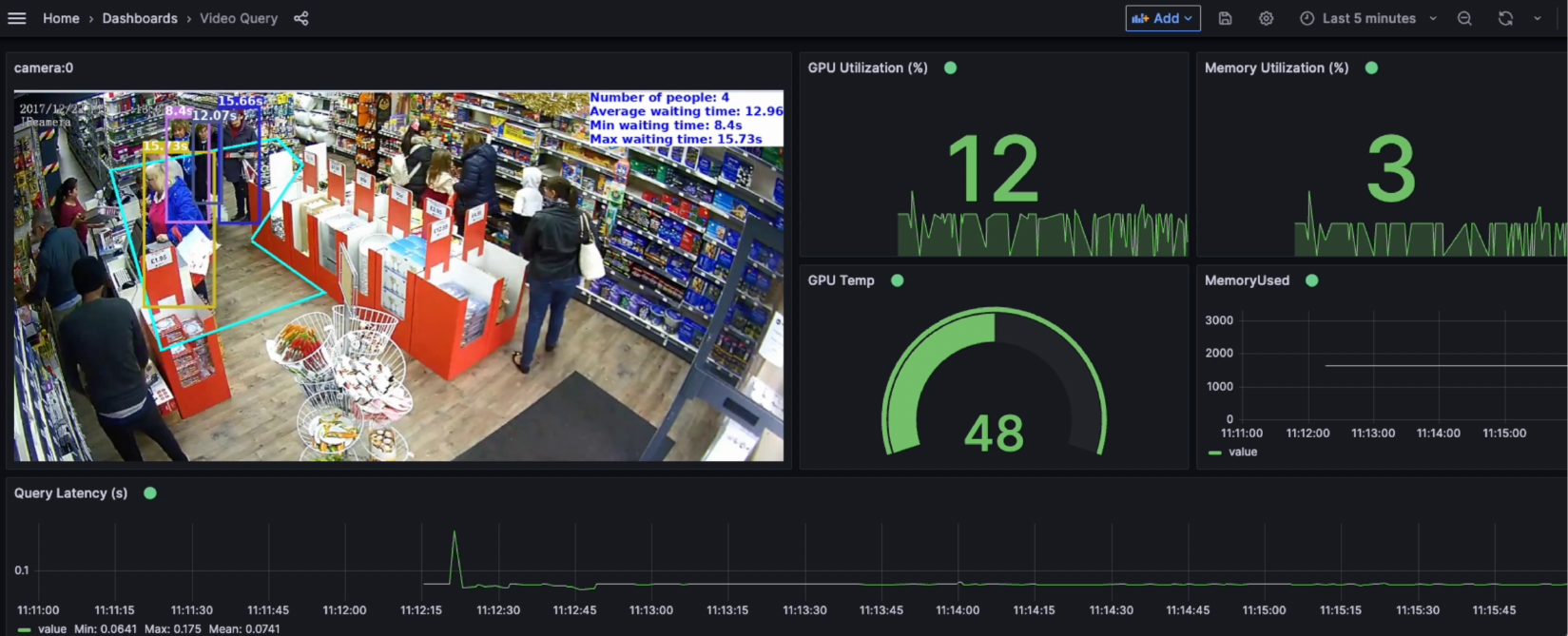}
         \footnotesize (b) Queue analysis
         \label{fig:dv-queue}
     \end{minipage}
     \vspace{-0.8em}
     \caption{Use cases of \tool with DeepVision.}
     \label{fig:dv}
\end{figure}

As of Oct 26, 2023, \tool has powered two critical user applications created for Cisco's Australian customers, \ie, loitering and queue analysis. Loitering alerting is essential for smart city safety. Queue analytics plays a vital role in retail store management. Figure~\ref{fig:dv} demonstrates how \tool integrates with DeepVision and delivers the results on DeepVision's Graphana-based dashboard.

\section{Related Work \label{sec:related}}

\MyPara{SQL-like languages.} Treating video frames as relational tables, prior work on video analytics often uses SQL-like languages to express video queries. Optasia~\cite{optasia}, BlazeIt~\cite{blazeit}, and EVA~\cite{eva} allow users to register NN UDFs and query frames whose properties satisfy given predicates. These frameworks, however, cannot express complex queries that involve multiple objects across multiple frames.

Miris~\cite{miris} and OTIF~\cite{otif} are frameworks specially designed for queries requiring cross-frame positional information (of bounding boxes), like a speeding car or a car moving toward a person. However, they cannot process queries on actions involving posture details (inside a bounding box), like people falling.  
SVQ++~\cite{svq++} and ~\cite{interactions} are frameworks targeting interactions between objects; they can express queries like "a person throw a ball". Zeus~\cite{zeus} is another specialized framework supporting queries on object interactions. 

EVA~\cite{eva} and VIVA~\cite{romero, viva} support queries on actions, but \emph{not} spatial and temporal relationships between video objects. Miris~\cite{miris} uses a customized \codeIn{Speed} UDF to express queries involving cross-frame information such as \codeIn{Speed(car) > 30 km/h}; SVQ++~\cite{svq++} requires a customized a \codeIn{THROW} UDF to express interactions between different objects (\eg, a person throws a ball). If the predefined UDFs and models cannot satisfy a user's need, the user must hand-write UDFs to deal with raw frames, a task that typically requires an in-depth understanding of the underlining systems. 
VIVA~\cite{romero} provides relation hints that allow users to express relationships between their own models and existing models in the framework.

\MyPara{Other query languages.}  Rekall~\cite{rekall} offers a Python library for video event specification, which allows users to iteratively specify and refine the temporal and spatial relationships of video segments. It cannot express queries including complex activities within or between objects, such as a person walking or riding a bicycle. 
Caesar~\cite{caesar} employs a text-based specification for users to express queries. It defines a vocabulary of basic actions such as "approach", "near", and "stop", and users can use these basic actions to compose a more complex action to use in a query. However, the vocabulary system is hard to extend and customize. Queries are constrained by the supported vocabulary.   

\MyPara{Specialized models.} Several video analytics systems use specialized models to either filter out frames to reduce the work of using expensive machine learning models or directly provide answers. Existing works mostly target specific query types such as selection (\eg, NoScope~\cite{noscope} and PP~\cite{probpredicate}) or aggregation (\eg, Blazelt~\cite{blazeit} and ABAE~\cite{abae}).

\MyPara{Model selection.} Other works propose using multiple models with different performance-accuracy trade-offs to adapt to the dynamic nature of the videos. TAHOMA~\cite{tahoma} and NoScope~\cite{noscope} employs a cascading approach that can skip compute-expensive models by answering the query with cheap models on easy frames. THIA\cite{thia} shares the same idea but uses a single model with multiple exit points to explore the performance-accuracy trade-off. Figo~\cite{figo}, instead of cascading the models, uses an ensemble of models with the same architecture but different sizes and dynamically selects the most cost-efficient model for the current video chunk.

\MyPara{Runtime optimizations.} Systems like Scanner~\cite{scanner}, Jellybean~\cite{jellybean}, VideoStorm~\cite{videostorm}, and LLAMA~\cite{llama} focus on optimizations of end-to-end  analytics pipelines via efficient use of heterogeneous hardware. These techniques do not focus on query expressions, and thus are also orthogonal to and can be integrated into \tool. 

\section{Conclusion}
\label{sec:discuss}

This paper presents \tool, a video-object-oriented system we built for easy expression and optimization of complex video analytics queries. 
\tool has an object-oriented frontend, enabling abstraction and modularity, as well as an extensible backend, allowing advanced developers to integrate optimizations into the \tool pipeline. We have open-sourced \tool, which is currently used in Cisco for a range of vision tasks.

\section*{Acknowledgments}
We thank the anonymous MLSys reviewers for
their thorough comments. This work is supported by NSF grants  CNS-1907352, CNS-2007737, CNS-2006437, CNS-2106838, CNS-2128653, CNS-2330831, and two grants awarded by Cisco Research.

\bibliography{main}
\bibliographystyle{mlsys2024}

\appendix
\section{Query Code Comparisons between \tool and EVA}
\label{sec:appendix-A}
\begin{figure*}[th!]
\begin{minipage}[t]{0.45\textwidth}
\includegraphics[width=\linewidth]{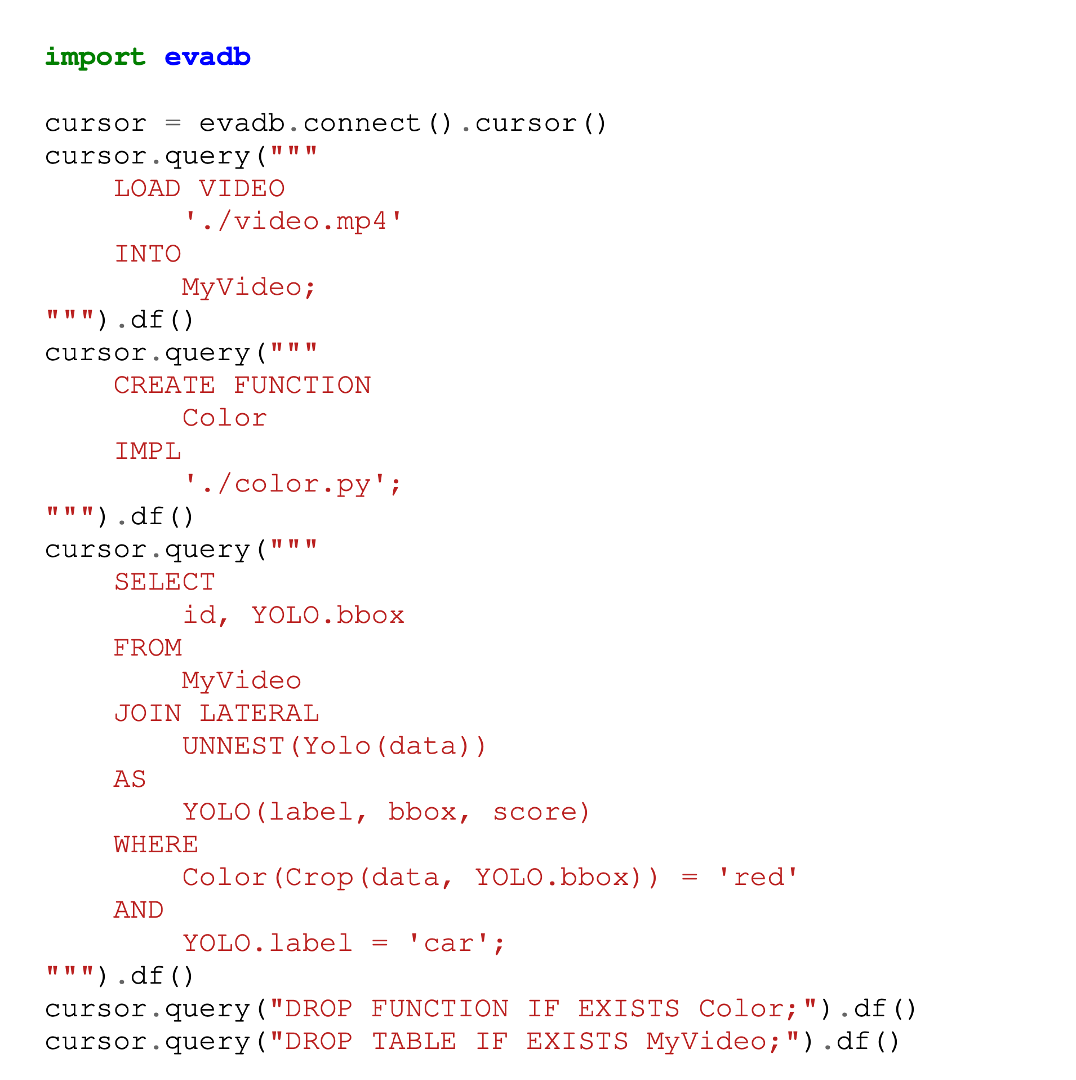}
\vspace{-1.5em}
\caption{EVA SQL expressions for querying red cars.
\label{fig:eva-red-car} \vspace{-1em}}
\end{minipage}
\hfill
\vspace{0.1in}
\begin{minipage}[t]{0.45\textwidth}

\includegraphics[width=0.95\linewidth]{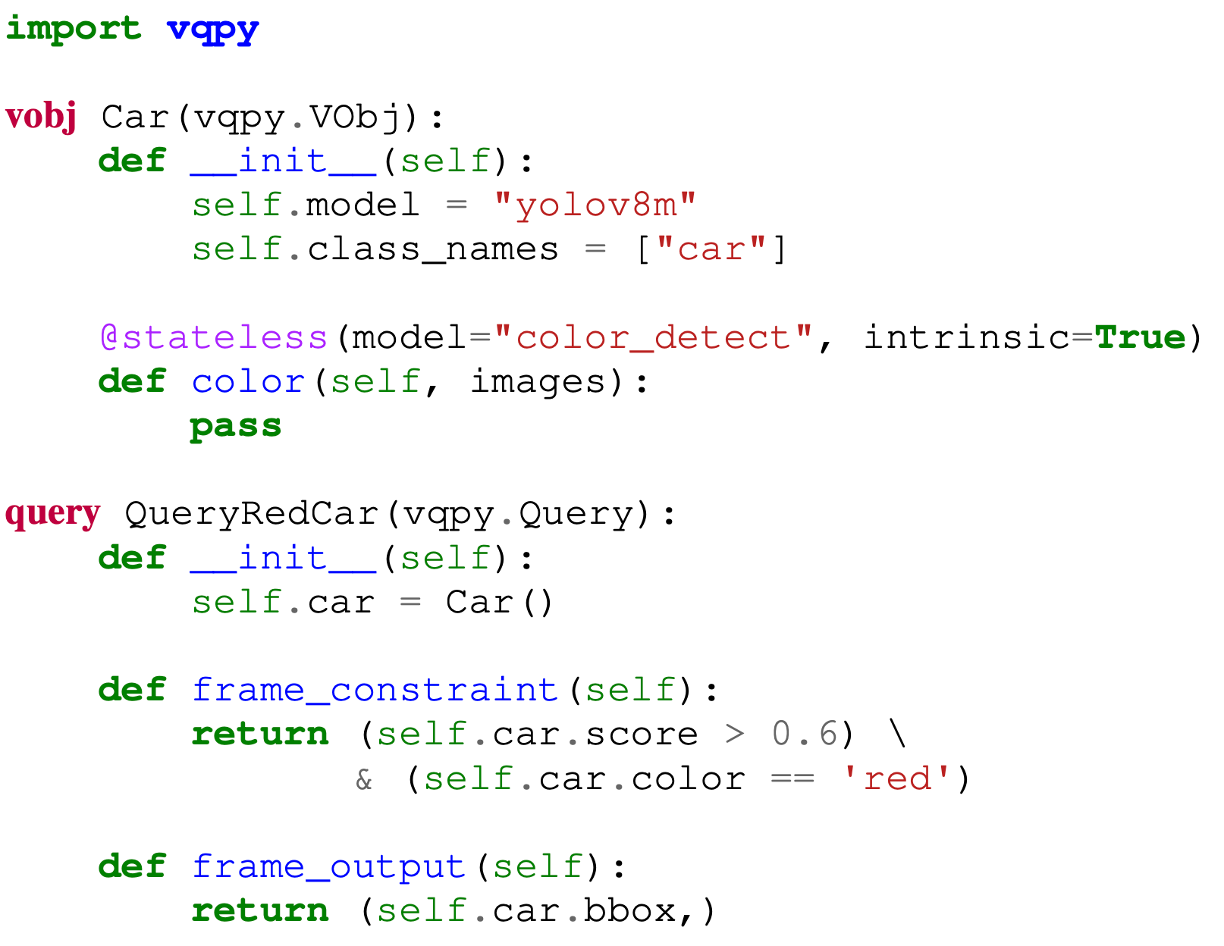}
\caption{VQPy expressions for querying red cars.
\label{fig:vqpy-red-car} \vspace{-1em}}
\end{minipage}

\end{figure*}

\begin{figure}[htbp]
\includegraphics[width=\linewidth]{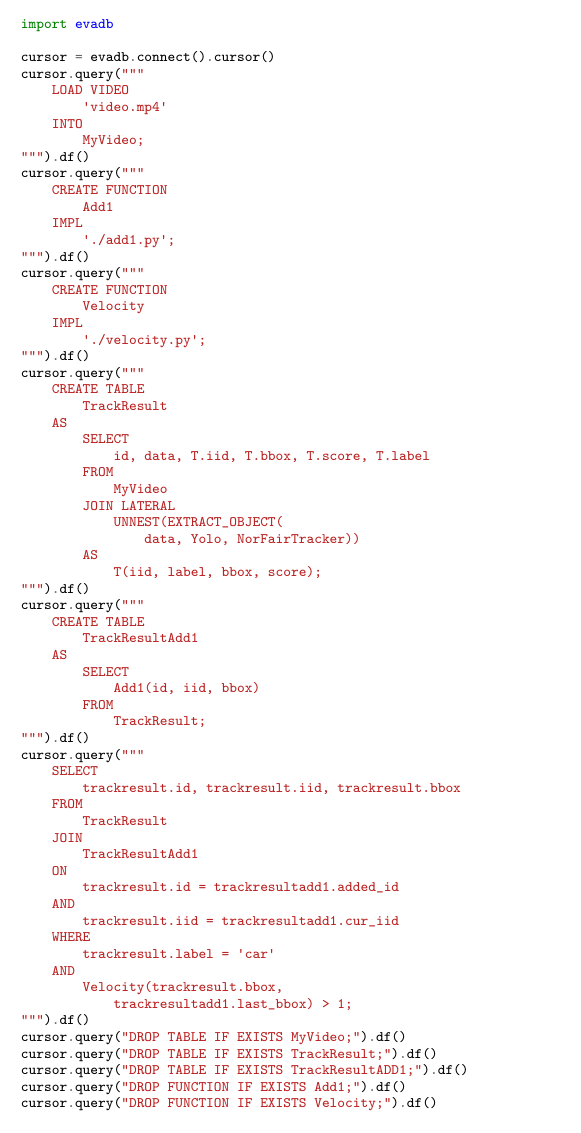}
\vspace{-1.5em}
\caption{EVA SQL expressions for querying speeding cars.
\label{fig:eva-speed-car} \vspace{-1em}}
\end{figure}

\begin{figure}[htbp]
\includegraphics[width=\linewidth]{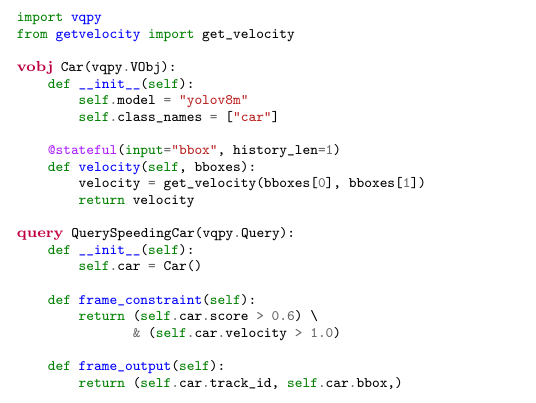}
\vspace{-1.5em}
\caption{VQPy expressions for querying speeding cars.
\label{fig:vqpy-speed-car} \vspace{-1em}}
\end{figure}

\begin{figure}[htbp]
\includegraphics[width=\linewidth]{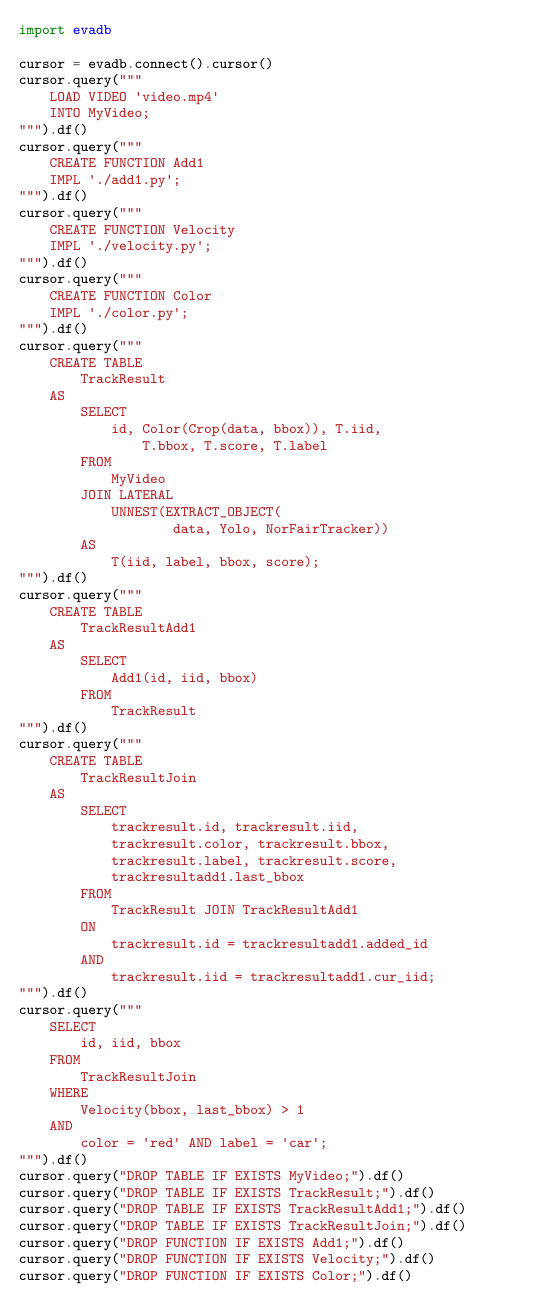}
\vspace{-1.5em}
\caption{EVA SQL expressions for querying red speeding cars.
\label{fig:eva-red-speed-car} \vspace{-1em}}
\end{figure}

\begin{figure}[htbp]
\includegraphics[width=\linewidth]{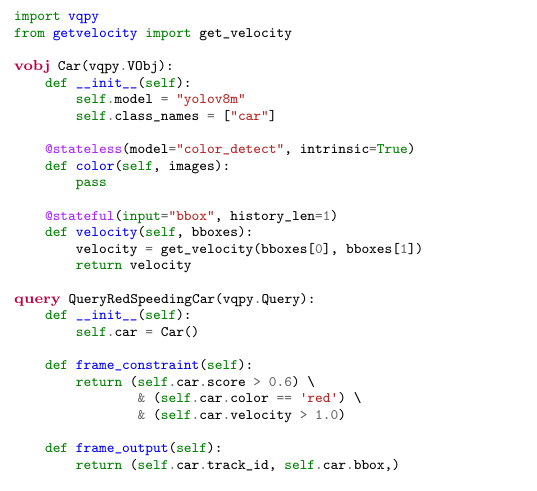}
\vspace{-1.5em}
\caption{VQPy expressions for querying red speeding cars.
\label{fig:vqpy-red-speed-car} \vspace{-1em}}
\end{figure}

This section compares three pairs of query programs (illustrated in Figure~\ref{fig:eva-red-car}-\ref{fig:vqpy-red-car}, Figure~\ref{fig:eva-speed-car}-\ref{fig:vqpy-speed-car}, and Figure~\ref{fig:eva-red-speed-car}-\ref{fig:vqpy-red-speed-car}) used to evaluate \tool and EVA (\S\ref{sec:eval-sql}). Programs in each pair are written in EVA and VQPy, respectively, to express the same query. As seen, the abstractions we provide in \tool make it much easier to develop and understand a query while developers must ``think like a table" when using a SQL-based framework, which is counter-intuitive and leads to code that is hard to understand and maintain. 

\end{document}